\documentclass[accepted]{article}

\usepackage{stfloats}
\usepackage{float}
\usepackage{subcaption}
\usepackage{microtype}
\usepackage{graphicx}
\usepackage{booktabs} 

\usepackage{hyperref}

\usepackage{icml2024}

\usepackage{amsmath}
\usepackage{amssymb}
\usepackage{mathtools}
\usepackage{amsthm}

\usepackage[capitalize,noabbrev]{cleveref}

\usepackage[algo2e,ruled,linesnumbered]{algorithm2e}
\theoremstyle{plain}

\theoremstyle{definition}

\theoremstyle{remark}

\usepackage[textsize=tiny]{todonotes}

\newcommand{\placeholder}{\mathord{\color{black!33}\bullet}}
\newcommand{\base}{\mathcal{L}_0}
\newcommand{\NguyenX}{$\base$}
\newcommand{\NguyenXY}{$\base \cup \{ x_2 \}$}
\newcommand{\NguyenXC}{$\base \cup \{ \textrm{const} \}$}
\newcommand{\NguyenXYCZ}{$\base \cup \{ x_2,\placeholder^{\placeholder},\textrm{const}\}$}
\newcommand{\NguyenXCZ}{$\base \cup \{\placeholder^{\placeholder},\textrm{const}\}$}
\newcommand{\NguyenXYC}{$\base \cup \{ x_2,  \textrm{const} \}$}
\newcommand{\NguyenXYZ}{$\base \cup \{ x_2,x_3  \textrm{const} \}$}
\newcommand{\NguyenXYZDC}{$\base \cup \{ x_2,x_3,x_4 , \textrm{const} \}$}
\newcommand{\NguyenXYtantanhZD}{$\base \cup \{ x_2,x_3,x_4, \textrm{const},\textrm{tan} ,\textrm{tanh}\}$}
\newcommand{\NguyenXall}{$\base \cup \{ x_2,x_3,x_4,x_5,x_6,x_7, x_8,x_9,x_{10} \}$}
\newcommand{\NguyenXYtanZD}{$\base \cup \{ x_2,x_3,x_4, \textrm{const},\textrm{tan} \}$}
\newcommand{\NguyenXYZDFCZ}{$\base \cup \{ x_2,x_3,x_4, x_5,\placeholder^{\placeholder},\textrm{const},\textrm{tan} \}$}

\newcommand{\NguyenXOne}{$\base \cup \{ 1 \}$}
\newcommand{\NguyenXFOne}{$\base \cup \{ -1 \}$}
\newcommand{\NguyenXYOne}{$\base \cup \{ x_2,1 \}$}
\newcommand{\Jin}{ $\base - \{ \log \} \cup \{\placeholder^2, \placeholder^3, x_2$ }
\newcommand{\Jintf}{ $\base - \{ \log \} \cup \{\placeholder^3, \placeholder^4, x_2, \textrm{const} \}$}
\newcommand{\nolog}{ $\base - \{ \log \}$}
\newcommand{\Jint}{ $\base - \{ \log \} \cup \{ \placeholder^3, x_2, \textrm{const} \}$}
\newcommand{\Nn}{$\base \cup \{\placeholder^4, \placeholder^3, x_2, \textrm{const} \}$}

\newcommand{\Keijzer}{$\{ +, \times, \div, \placeholder^{-1}, -\placeholder, \sqrt{\placeholder}, x \}$}

\newcommand{\VladislavlevaB}{$\{ +, -, \times, \div, \exp, e^{-\placeholder}, \placeholder^2, x, x_2 \}$}
\icmltitlerunning{Submission and Formatting Instructions for ICML 2024}

\begin{document}

\twocolumn[
\icmltitle{Discovering Mathematical Formulas from Data via GPT-guided Monte Carlo Tree Search}



\icmlsetsymbol{equal}{*}

\begin{icmlauthorlist}
\icmlauthor{Yanjie Li}{yyy,sch}
\icmlauthor{Weijun Li}{equal,yyy,sch,comp}
\icmlauthor{Lina Yu}{yyy}
\icmlauthor{Min Wu}{yyy}
\icmlauthor{Jingyi Liu}{yyy}
\icmlauthor{Wenqiang Li}{yyy,sch}
\icmlauthor{Meilan Hao}{yyy}
\icmlauthor{Shu Wei}{yyy}
\icmlauthor{YuSong Deng}{yyy}

\end{icmlauthorlist}

\icmlaffiliation{yyy}{ANNLAB, Institute of Semiconductors, Chinese Academy of Sciences, Beijing, China}
\icmlaffiliation{comp}{School of Electronic, Electrical and Communication Engineering, University of Chinese Academy of Sciences, Beijing, China}
\icmlaffiliation{sch}{School of Integrated Circuits, University of Chinese Academy of Sciences, Beijing, China}

\icmlcorrespondingauthor{Weijun Li}{liyanjie@semi.ac.cn}

\icmlkeywords{Machine Learning, ICML}
\vskip 0.3in
]



\printAffiliationsAndNotice{\icmlEqualContribution} 

\begin{abstract}
Finding a concise and interpretable mathematical formula that accurately describes the relationship between each variable and the predicted value in the data is a crucial task in scientific research, as well as a significant challenge in artificial intelligence. This problem is referred to as symbolic regression, which is an NP-hard problem. In the previous year, a novel symbolic regression methodology utilizing Monte Carlo Tree Search (MCTS) was advanced, achieving state-of-the-art results on a diverse range of datasets. although this algorithm has shown considerable improvement in recovering target expressions compared to previous methods, the lack of guidance during the MCTS process severely hampers its search efficiency. Recently, some algorithms have added a pre-trained policy network to guide the search of MCTS, but the pre-trained policy network generalizes poorly. To optimize the trade-off between efficiency and versatility, we introduce SR-GPT, a novel algorithm for symbolic regression that integrates Monte Carlo Tree Search (MCTS) with a Generative Pre-Trained Transformer (GPT). By using GPT to guide the MCTS, the search efficiency of MCTS is significantly improved. Next, we utilize the MCTS results to further refine the GPT, enhancing its capabilities and providing more accurate guidance for the MCTS. MCTS and GPT are coupled together and optimize each other until the target expression is successfully determined. We conducted extensive evaluations of SR-GPT using 222 expressions sourced from over 10 different symbolic regression datasets. The experimental results demonstrate that SR-GPT outperforms existing state-of-the-art algorithms in accurately recovering symbolic expressions both with and without added noise.
\end{abstract}

\section{Introduction}
The ultimate goal of natural scientific research is to discover a concise and clear mathematical expression from a set of data, which can describe the relationship between variables in the data and reflect the objective laws of the physical world. The goal of symbolic regression is to use observed data to search for an expression that can fit the data well. Specifically, if there is a set of data $[x_1,x_2,...,x_m,y]$ where $x_i\in \mathbb{R}^n$ and $y\in\ \mathbb{R}$, the purpose of symbolic regression is to discover a mathematical expression $f(x_1,x_2,...,x_m)$ through certain methods so that $f$ can fit the data $y$ well. The resulting expression can not only fit the data $y$ well but also be interpretable. We can utilize the properties of the basic operators in the expression $f$ to analyze the relationship between the feature variables $[x_1,x_2,...,x_m]$ and $y$ in the data. 
In recent years, deep learning has penetrated various areas of our lives\citep{chen2022recent,choudhary2022recent,zhang2021application}. People from many fields, from physics to life sciences, are using neural networks to fit their data. Although artificial neural networks can fit the data well, the result obtained by the neural network fitting is a "black box"\citep{petch2022opening}, which is not interpretable and analyzable. On the contrary, the result of symbolic regression is a clear, interpretable, and analyzable mathematical expression. For example, in the physical formula $\mathcal{P = FV}$, we can easily analyze that, given a constant power $\mathcal{P}$, to obtain a larger force $\mathcal{F}$, we must decrease the velocity $\mathcal{V}$. This is also why vehicles slow down when going uphill in real-life scenarios. However, a black-box model obtained from a neural network cannot intuitively derive such useful conclusions. \\  
Mathematical expressions are composed of basic operators, and any expression can be expressed as an expression binary tree. If we expand the expression binary tree in the order of preorder traversal, we can obtain an ordered and discrete sequence of operators. So essentially symbolic regression can be regarded as a combinatorial optimization problem \cite{karimi2022machine}. This is an NP-hard problem \citep{huynh2022discovery}. Traditional approaches to symbolic regression typically utilize evolutionary algorithms, especially genetic programming (GP)\citep{koza1992evolution,schmidt2009distilling,haarnoja2018soft}. In GP-based symbolic regression, the algorithm begins by initializing a population of expressions. The individuals in the population undergo crossover and mutation operations to simulate human evolution. Finally, the best individuals are selected to form the next generation population. While GP can be effective, it is also known to scale poorly to larger problems and exhibit high sensitivity to hyperparameters. \\
Symbolic Physics Learner (SPL)\cite{sun2022symbolic}, published this year, employs MCTS to address symbolic regression problems. In SPL, MCTS iteratively executes four steps (selection, expansion, simulation, and backtracking) to progressively uncover the optimal search path. This paradigm effectively tackles symbolic regression problems and demonstrates remarkable performance; Appendix \ref{sec:MCTS} delineates a succinct example of employing MCTS to tackle symbolic regression. Nonetheless, employing a strategy of completely random selection during the initial expansion and the subsequent simulation phases, in the absence of a guiding mechanism, leads to suboptimal search efficacy in SPL. The DGSR-MCTS\cite{DGSR-MCTS} and TPSR\cite{sr-plan} integrate a pre-trained policy network to facilitate the search mechanism of the MCTS. However, this pre-trained model demonstrates limited generalizability, exhibiting satisfactory performance only on data configurations encountered during its training phase. For instance, should the training be confined to a symbolic repository encompassing $[+, sin, cos, x]$, any subsequent attempts to extend the repository to incorporate additional symbols such as $[+, -, sin, cos, exp, x]$ results in suboptimal performance, or may altogether fail to function. Furthermore, the model's efficacy is significantly compromised when exposed to scenarios deviating from its training conditions; a model trained exclusively with two variables, $x_1$ and $x_2$, is markedly less competent when evaluated on datasets featuring three variables. Even when the model samples X within the interval [-2, 2] during training, its performance is greatly compromised when we sample X within the range of [-4, 4] for testing. Moreover, the anti-noise ability of these algorithms is relatively poor. It's hard to apply to real-world problems. In light of these considerations, both the generalization and search efficiency of the algorithm must be addressed. To this end, we introduce an innovative algorithm, termed SR-GPT, designed to reconcile the two.\\
We propose $SR-GPT$, a novel symbolic regression framework based on GPT and MCTS. In $SR-GPT$, we integrate the policy network and value network into one. We employ a GPT to generate the probability $p$ of selecting each symbol and the value $v$ of the current state, which guides the MCTS during the expansion and simulation phases. Before the self-search generates a symbol, multiple simulations are conducted. Once the self-search generates a complete expression. we calculate the reward $z$ using the reward function and perform backpropagation. Furthermore, we collect the node information selected during the self-search procedure to train the GPT. Specifically, in the self-search phase we get a preorder traversal of the complete expression $[a_1,a_2,a_3,...] $, we have a probability $\pi_t$ when we choose each symbol $a_t$, as shown in Fig \ref{fig1}a. After obtaining the complete expression, we calculate the reward value $z$. Therefore, we can use the sequence $[a_1,a_2,a_3,...] $ as GPT input, let GPT output $[[\pi_1,z],[\pi_2,z],...] $. For example, if we input $[a_1]$, GPT outputs $[\pi_2,z]$. Input $[a_1,a_2]$, GPT output $[\pi_3,z]$... Every time we get a complete expression we get a training data pair. We'll store it in a queue of size 1000 and randomly sample batch size pairs from the queue each time we train. If the queue is full, the old one goes out and the new one comes in. Using these data, we can train the GPT to become even more powerful, which can better guide the MCTS. 
We summarize our contributions as follows:  
\begin{itemize}
\item We propose a novel symbol regression model, called SR-GPT, which cleverly implements the mutual coupling between GPT and MCTS. And outperforms several baselines on a series of benchmark problems.
\item  We improve the loss function (used to train GPT) to encourage the GPT to produce a probability distribution with lower information entropy, thereby avoiding situations where each symbol is predicted with a similar probability. Improved the search efficiency of the algorithm.
\item We propose a new loss function (Used to calculate rewards) called $S_{NRMSE}$, which effectively addresses the issue of variable omission in multivariate regression problems.
\end{itemize}
\section{Related Work}
\textbf{Deep learning for symbolic regression}.  
Recently, many algorithms have been developed to apply deep learning to symbol regression, achieving promising results. EQL\citep{eql1,eql2} replaces the activation function of neural networks with basic operators such as $[+,-,*,/, sin,...]$, Afterwards, parameter sparsification is conducted to eliminate redundant connections and extract expressions from the network.  
AI Feynman consists of two versions, AI Feynman 1.0\citep{aif1} and AI Feynman 2.0\citep{aif2}, both of which aim to simplify complex problems. In AI Feynman 1.0, a neural network is first trained to fit data and then used to discover some properties, such as additivity separability, Then, these properties are used to break down a complex problem into several simpler ones. The limitation is that AI Feynman 1.0 applies only a limited number of properties and achieves better performance only in the domain of physics expressions.  Building upon the foundation of AI Feynman 1.0, version 2.0 proposes more properties, expanding the algorithm's applicability to any field. 
The NeSymRes\citep{biggio2021neural} algorithm treats symbolic regression as a machine translation problem. This algorithm trains a transformer model with a large amount of data and then uses the model with beam search to generate expressions. This article\citep{l1} replaces the feature extraction module of NeSymRes with the point cloud feature extraction algorithm pointMLP.   \\
\textbf{Genetic programming for symbolic regression}.
 Genetic Algorithm (GA)\citep{gp2,gp3,gp4} is a classic optimization algorithm that simulates the evolution of human history. The symbolic regression algorithm based on GA is Genetic programming (GP). GP first initializes a population of expression binary trees. Then, a simulation of human ``evolution'' is carried out through means such as crossover and mutation. Finally, excellent individuals are selected as the next-generation population through fitness selection. This process is repeated. In addition, there are many algorithms that have been developed by improving upon the GP algorithm\citep{gp2,gp3,gp4}.\\
\textbf{Reinforencement learning for symbolic regression}.
DSR\citep{dsr} and DSO \citep{dso} are two excellent symbolic regression algorithms based on deep reinforcement learning. Among them, DSR defines Recurrent Neural Networks (RNN\citep{rnn}, LSTM\citep{lstm}, GRU \citep{gru}, etc) as the policy network. The network takes the parent and sibling nodes of the node to be generated as inputs and outputs the probability of selecting each symbol. Then, multiple expressions are sampled from the policy network. Finally, the reward values are calculated, and the policy network parameters are updated through policy gradient to enable the policy network to generate better expressions. DSO introduces the GP based on DSR. In DSO, the expressions sampled from the policy network are used as the initial population for the GP. Then the GP-evolved expressions and original expressions are combined to update the policy network to provide higher-quality initial populations for GP.  SPL\citep{sun2022symbolic} applies the successful conventional MCTS to the symbolic regression field. The algorithm selects a good sequence of expressions by repeating four steps: selection, expansion, simulation, and backpropagation. Both DGSR-MCTS\citep{DGSR-MCTS} and TPSR\citep{sr-plan} introduce a pre-trained model to guide the MCTS search process.
\section{Modeling}
\begin{figure*} [htbp]
    \centering
	  \subfloat{
       \hspace{0.3cm}
       \includegraphics[width=12.8cm,height=7.2cm]{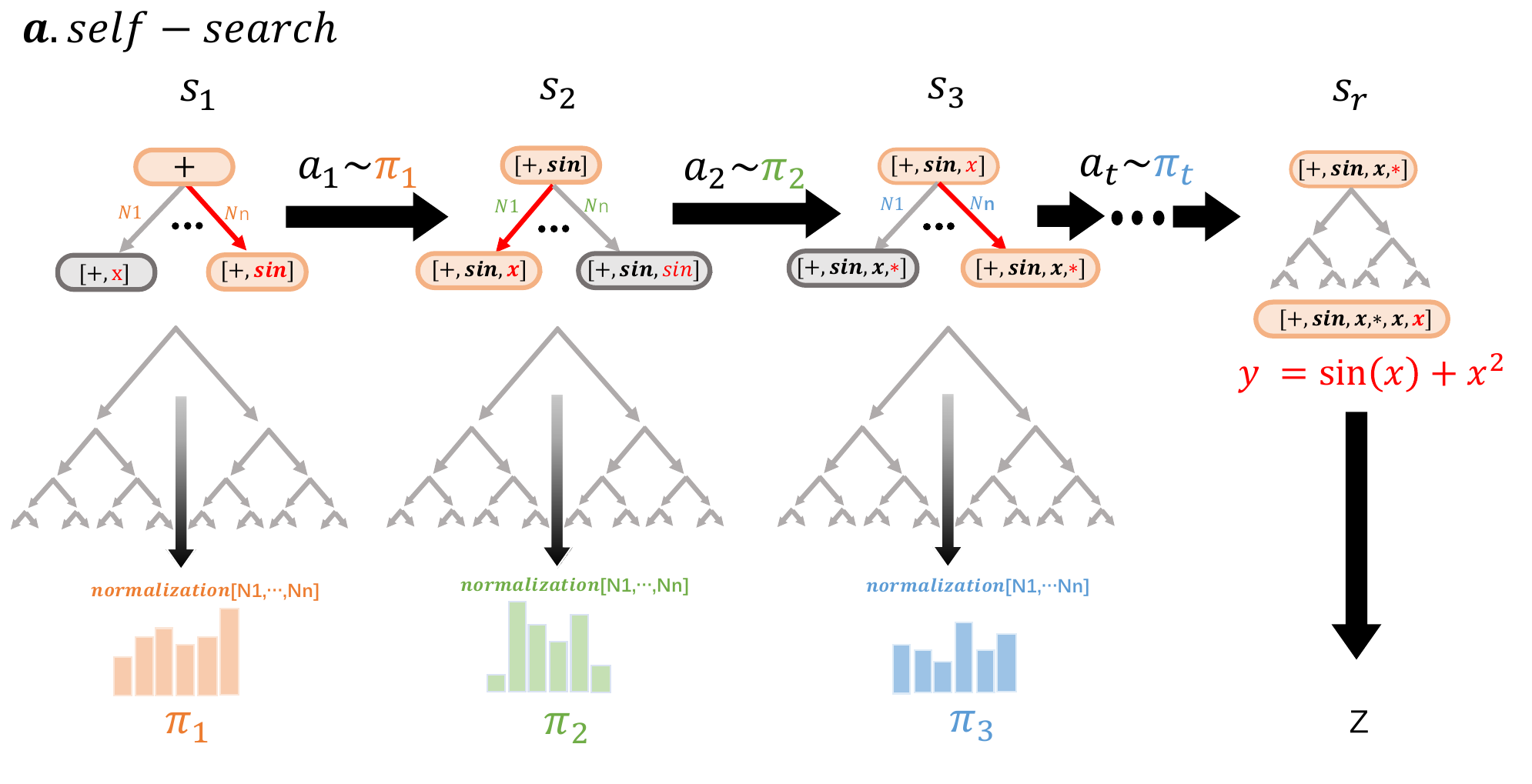} \label{fig1a}}
        \hfill
	  \subfloat{ \includegraphics[width=12.4cm,height=7.4cm]{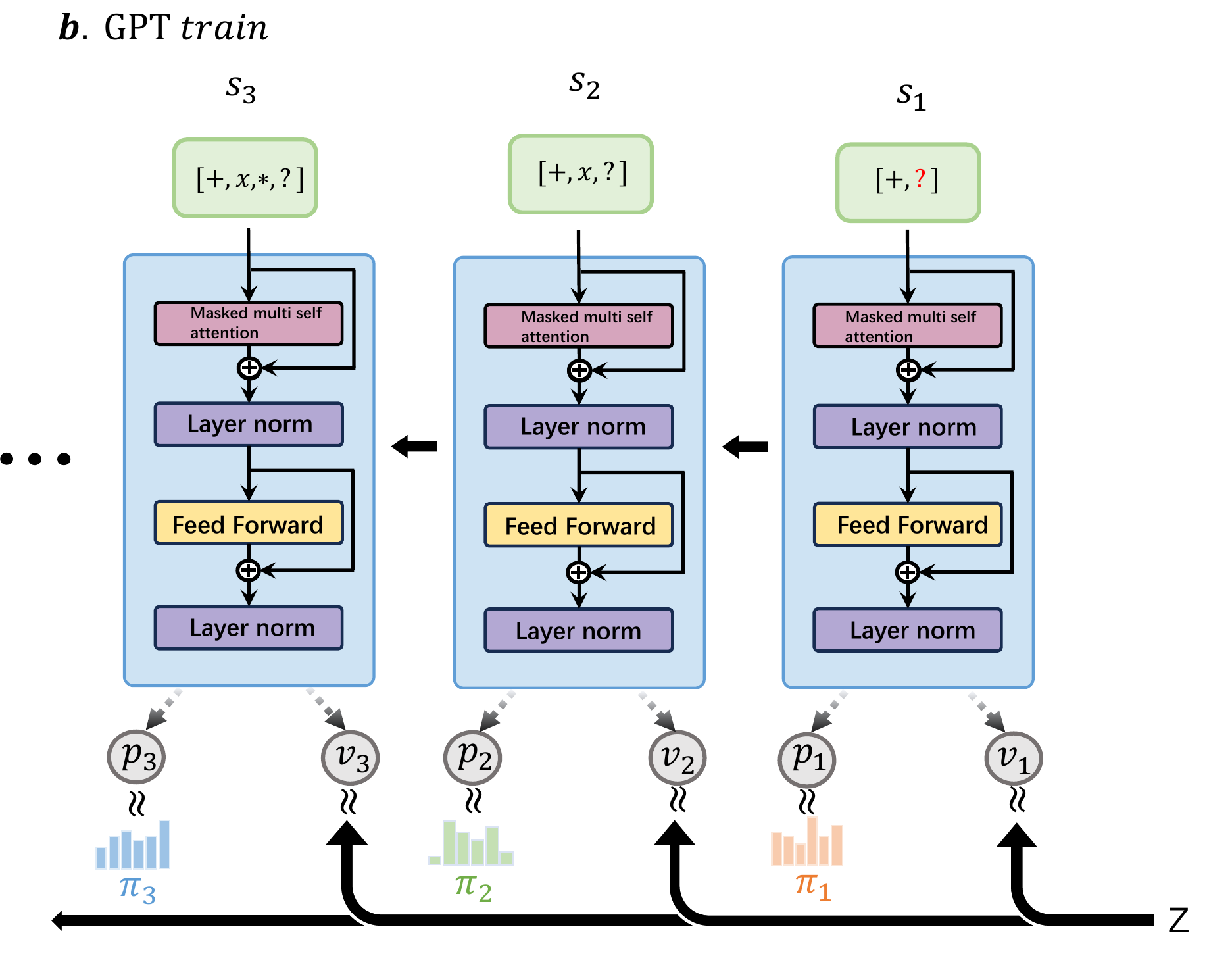}\label{fig1b}}  
    	  \caption{\textbf{a. self-search.} During the self-search phase, the operators with the highest probabilities(max($\pi$)) are selected sequentially for each state $s_1, s_2, ..., s_t$. For each state $s_t$, multiple simulations are carried out, and during each simulation, an MCTS $\alpha(\theta)$ guided by the latest neural network $\mathcal{N}_{\theta}$ is executed (as depicted in Fig \ref{fig2}). Finally, by normalizing the number of times each child state of the current state has been visited, we can determine the selection probability $a_{t}\sim\pi_{t})$ for each state. Finally, when the expression is complete at time $s_r$, we calculate the reward value $z$ and perform backpropagation. \textbf{b. GPT training. } In SR-GPT, the GPT is designed to take the state $s_t$ as input, which is then passed through an GPT  with parameters $\theta$. The output comprises a vector $p_t$ and a scalar value $v_t$, where $p_t$ represents a probability distribution over candidate symbols and $v_t$ represents the possible reward value after generating a complete expression starting from the current state $s_t$. During training, the GPT parameters, $\theta$, are updated to maximize the similarity between the policy vector, $p_t$, and the search probabilities, $\pi_t$, while minimizing the difference in the predicted reward $v_t$ and the actual reward $z$. The new parameters are used in the next iteration of self-search as in Fig (a).
   }
	  \label{fig1} 
\end{figure*}
\begin{figure*}[htbp]
\centering
\includegraphics[width=150mm]{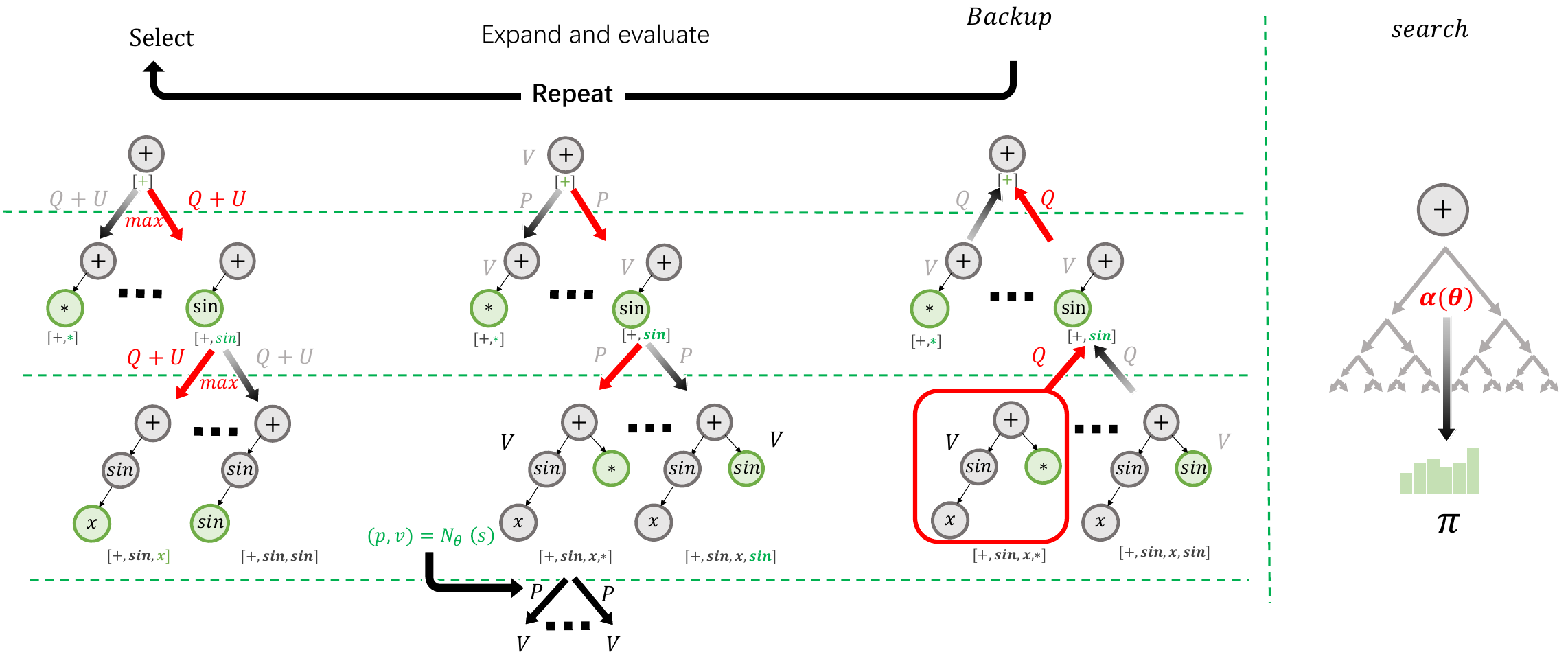}
\caption{
\textbf{MCTS in SR-GPT}. \textbf{Select, } Starting from the root node, the child node with the largest UCT value is selected recursively, where $\mathcal{UCT}=\mathcal{Q}(s, a)+\mathcal{U}(s, a)$, and finally a leaf node is reached. 
\textbf{Expand,} If the current leaf is not a terminal node, then create multiple (number of symbols) child nodes, and select the one with the greatest probability $p(s, a)$ to expand, where $p(s, \bullet) = \mathcal{N}_{\theta}(s)$, and the $p(s, a)$ values are stored in the corresponding new child nodes. \textbf{evaluate, } Use GPT to calculate the $V$ of the new extension node.
\textbf{backpropagation,} After each simulation, backpropagation is performed, where $\mathcal{N}_{visit}$ of all visited nodes during the simulation is incremented by one. Additionally, Action value $Q$ is updated to track the mean of all evaluations V in the subtree below that node\ref{e2}.
\textbf{Search, } The probability $\pi$ of selecting each symbol can be determined during the self-search phase after the completion of the simulation phase. Herein, $\pi$ is directly proportional to $(\mathcal{N}_{visit})^{\frac{1}{\tau}}$, where $\mathcal{N}_{visit}$ is the visit count of each move from the root state and $\tau$ is a parameter controlling temperature.
}
\label{fig2}
\end{figure*}
The symbol library contains a series of basic operators that can be flexibly combined into various mathematical expressions. Our symbol library includes five binary operators [$+$, $-$, $\times$, $\div$, $\bullet^{\bullet}$], five unary operators [$\sin$, $\cos$, $\mathrm{exp}$, $\sqrt{\phantom{x}}$, $\ln$], multiple variable operators [$x_1, x_2, \dots, x_n$], as well as constant placeholders [$c$]. Guided by an 
GPT, the SR-GPT executes MCTS, generating operators for the expression in the order of preorder traversal of the expression's binary tree.\\  
SR-GPT uses an GPT $\mathcal{N}_{\theta}$ with parameter $\theta$. This GPT combines the roles of both the policy network and value network into a single architecture. $\mathcal{N}_{\theta}$ takes as an input the parent node and sibling nodes of the node to be predicted and outputs the probability $p$ of selecting each operator and the current state value $v$, $(p,v)=\mathcal{N}_{\theta}(s)$. Where the state value $v$ can be seen as the degree to which continuing downward selection from the current node can eventually lead to the optimal expression.\\      
The GPT in SR-GPT is trained from the task of self-search by a reinforcement learning algorithm. In self-search, before selecting a new symbol, we will conduct several simulations under the guidance of GPT $\mathcal{N}_{\theta}$. After several MCTS simulations, the number of times $N_i$ each possible child node is selected is obtained, and then $[N_1,N_2,...N_n]$ is normalized to obtain $\pi$. Then, based on the probability $\pi$, the program is instructed to actually choose a symbol. These search probabilities usually select much stronger moves than the raw move probabilities $p$ of the GPT $\mathcal{N}_{\theta}(s)$; MCTS may therefore be viewed as a powerful policy improvement operator. When we have a completed expression, we will calculate the reward $z$ through the reward function, $\mathcal{R}(s)=1/(1+S_{NRMSE})$, where $S_{NRMSE}$ is shown in expression \ref{e6}. Repeat the process until you get the target expression. the GPT's parameters are updated to make the operator's probabilities and value $(p, v) = \mathcal{N}_{\theta}$ more closely match the improved search probabilities and self-search reward $(\pi, z)$; these new parameters are used in the next iteration of self-search to make the search even stronger. Figure \ref{fig1} a illustrates the self-search training process.\\       
As shown in Figure\ref{fig2}, MCTS is simulated several times under the guidance of an GPT. During the simulation, each node stores three kinds of information, prior probability $\mathcal{P}(s, a)$, a visit count $\mathcal{N}(s, a)$, and an action value $\mathcal{Q}(s, a)$. In the simulation stage, each simulation starts from the root node, and the node with the largest $\mathcal{UCT}$\citep{uct} value among the child nodes is selected successively. The $\mathcal{UCT}$ expression is shown in expression below \ref{e1}:

\begin{equation}
\label{e1}
\displaystyle \mathcal{\mathcal{UCT}}(s,a) = \mathcal{Q}(s,a) + \mathcal{U}(s,a) 
\end{equation}
 Here Action value $\mathcal{Q}(s, a)$ \citep{agz}is the cumulative average reward value of the current state, and the expression is \ref{e2}:
\begin{equation}
\label{e2}
\displaystyle \mathcal{Q}(s, a)=\frac{1}{\mathcal{N}(s, a)}\sum _{s^{'}|s,a\rightarrow s^{'}}  \mathcal{V}(s^{'}) 
\end{equation}
And $s, a\rightarrow s^{'}$ indicates that a simulation eventually reached $s^{'}$ after taking move $a$ from position $s$. where  $\mathcal{U}(s, a)\propto p(s, a)/(1+\mathcal{N}(s, a)$, The specific expression is \ref{e3}: 
\begin{equation}
\label{e3}
\displaystyle \mathcal{U}(s, a) = c_{puct} * \mathcal{P}(s, a) * \frac{\sqrt{\mathcal{N}(s)}}{1 + \mathcal{N}(s, a)}
\end{equation}
When a leaf node $s^{'}$ is encountered during the simulation, if the leaf node is not a terminal node, this leaf position is expanded and evaluated only once by the network to generate both prior probabilities and evaluation, $(\mathcal{P}(s^{'},\bullet), \mathcal{V}(s^{'})) = \mathcal{N}_{\theta}(s^{'})$. Finally, a backpropagation step is performed to update the $ counter$, action value $\mathcal{Q}(s, a)$, visit count $\mathcal{N}(s, a)$, and $length$ for the nodes traversed during the simulation.
The self-search process of SR-GPT can be perceived as a symbolic selection process where symbols are chosen through a greedy search algorithm guided by a probability $\pi$. The symbol with the maximum $\pi$ value is selected as a new symbol. In addition, $\pi$ is proportional to the number of visits of each child node after multiple simulations. Here $\pi_a \propto \mathcal{N}(s,a)^{\frac{1}{\tau}}$, and its precise computational expression is as follows \ref{e4}.
\begin{equation}
\label{e4}
\displaystyle \pi_{a_i} = \frac{log(\mathcal{N}(s,a_i)^{\frac{1}{\tau}})}{\sum_{i=0}^{n}log(\mathcal{N}(s,a_i)^{\frac{1}{\tau}})}
\end{equation}  
The training of GPT is a regression problem. Firstly, we initialize the state as $s_{0}$, and the parameters of the GPT are randomly initialized as $\theta_{0}$. During the self-search phase, before making an actual selection of a symbol, we conduct multiple MCTS simulations under the guidance of the GPT, with the current state $s_{0}$ as the input. Then, we obtain the probability $\pi_{0}$ and select symbol $a_{0}$ to move into state $s_{1}$ under the guidance of $\pi_{0}$. Train and update the parameters of the GPT using the data collected during the self-search phase to obtain $\theta_{1}$. Repeat the above process until a complete expression is obtained or the maximum length is reached. Once a complete expression is obtained, we compute the final reward $z$ and perform backpropagation. Each traversed node during the self-search process can contribute to training data $(s_{t}, \pi_{t}, z_{t})$, where $z_{t} = z$. Figure \ref{fig1} b illustrates that the GPT neural network is trained by randomly selecting a portion of the previously collected training data $(s,\pi,z)$, which generates a new set of parameters designated as $\theta_{new}$. The 
GPT $(p,v)=\mathcal{N}_{\theta new}(s)$ is adjusted to minimize the error between the predicted value $v$ and the reward $z$ and to maximize the similarity of the neural network predicts probabilities $p$ to the search probabilities $\pi$.   
Specifically, In this study, the parameters $\theta$ of the GPT are updated via the gradient descent method. Expression \ref{e5} shows the loss function to be optimized. It is noteworthy that we incorporate the information entropy of the predicted probability $p$ as a part of the loss function.  
\begin{equation}
\label{e5}
\displaystyle \mathcal{L} = (z-v)^2 - \pi^Tlog{p} - p^Tlog{p}+ \xi ||\theta||^2 
\end{equation}
Among them, the first term $(z-v)^2$ makes the predicted value $v$ and the true value $z$ as close as possible. The second term, $-\pi^T \log{p}$, minimizes the difference between predicted probability $p$ and $\pi$. The third term, $-p^T \log{p}$\citep{entropy}, the information entropy loss which maximizes the difference between the predicted probability of each symbol so that the probabilities are concentrated on a single symbol rather than being almost equal for every symbol. The final term, $\xi ||\theta||^2$, is the $L_2$ \citep{l2p} regularization penalty to prevent overfitting of the neural network, where $\xi $ is a parameter that controls the strength of $L_2$ weight regularization. The pseudocode for SR-GPT is shown in Algorithm \ref{algorithm1}\\
\textbf{Constant optimization}. If a constant placeholder "C" appears in the sampled expression, we will use nonlinear optimization algorithms such as BFGS \citep{bfgs} or L-BFGS-B \citep{lbfgsb} to optimize the constants. For example, for the expression $2.2*sin(x)+1.3$, we might search for an expression of the form $C * sin(x) + C$, and then perform constant optimization using BFGS to obtain the original expression.\\
\textbf{Loss function : $S_{NRMSE}$}. Traditional loss functions only calculate loss by evaluating the match between predicted value $\hat{y}$ and actual value $y$. The expression for the loss is $NRMSE = \frac{1}{\sigma}\sqrt{\frac{1}{N}\sum_{i=1}^{N}({y_i-\hat{y}_i})^2}$\citep{nrmse}. When dealing with multivariate problems, especially when one variable is numerically much smaller than the others, or when two variables are highly correlated, using NRMSE as a loss metric may easily result in variable omission. To tackle this challenge, we propose a new loss function called `$S_{NRMSE}$', with the following expression:\ref{e6}\\
\begin{equation}
\label{e6}
\begin{aligned}
\mathcal{S_{NRMSE}} &= \text{NRMSE}(y,\hat{y}) + \sum_{j=1}^{m}\text{NRMSE}(x_j,\hat{x}_j)\\
&= \frac{1}{\sigma_{y}}\sqrt{\frac{1}{\mathcal{N}}\sum_{i=1}^{\mathcal{N}}(y_i-\hat{y}_i)^2} \\
& \quad + \lambda \sum_{j=1}^{m}\frac{1}{\sigma_{x_j}}\sqrt{\frac{1}{\mathcal{N}}\sum_{i=1}^{\mathcal{N}}(x_{ji}-\hat{x}_{ji})^2}
\end{aligned}
\end{equation}

Here, $\mathcal{N}$ represents the sample size, $m$ represents the number of variables, and $x_{ji}$ refers to the $i^{th}$ variable of the $j^{th}$ sample. And $\hat{x_{ji}}$ denotes the predicted value of $x_{ji}$. Specifically, if the predicted expression contains the variable $x_j$ then $\hat{x_j}=x_j$. If the predicted expression is missing the variable $x_j$, then $\hat{x_j}=0$. $\lambda$ belongs to the interval $[0,1]$, indicating the importance of $\mathcal{X}$ in the loss function.
\\
\textbf{Reward function}. When a complete expression is obtained during the self-search phase, we need to use a reward function to evaluate its quality. Firstly, the reward function must meet the requirement that the smaller loss, the larger reward. Secondly, to make the learning process more stable, the reward function should be designed to have values that range between $[0,1]$. Therefore, we define the reward function as  
$Reward(s) = \frac{1}{1 + S_{NRMSE}} $\\
\textbf{Constraining the search space}. To ensure the smooth operation of the algorithm and enhance its efficiency and performance. We have imposed some reasonable constraints on the search space of the algorithm. (1), We have placed a limit on the length of the generated expression. (2), Operators that are inverse functions cannot occur consecutively (for example, the operator $exp(log(x))$ is not allowed). (3), We have disallowed the repeated nesting of $sin$ and $cos$  because, in real-life applications, expressions with repeated nesting of trigonometric functions are rare (e.g. $ sin(cos(x)) $ is not allowed). (4), Some illegal expressions are not allowed to be generated. For example, $log(\tau)$, where $\tau$ should be greater than zero, so in the process of expression generation, functions that have negative values like $sin$, $cos$, and so on can't come right after $log$.\\
\textbf{Expression generation termination}. Introduce two variables, $counter$ and $arity$\citep{dsr}, where the $counter$ is initialized to 1. Binary functions, such as $[+,-,\times,\div...]$, have an $arity$ of 2, while unary functions, such as $[sin, cos,exp, ln...]$, have an $arity$ of 1. Variables, $[x_1,x_2,...,x_n]$, and constants placeholders $[c]$ have an $arity$ of 0.
During the process of generating expressions through pre-order traversal, the current $counter$ of the symbol sequence is calculated in real-time using the expression $counter = counter + arity(a) -1$. If the $counter$ of the current sequence is 0, it indicates that a complete expression has been sampled, and the sampling is stopped. 
\begin{table*}[t]
\renewcommand{\arraystretch}{1.1}
\centering
\caption{Recovery rate comparison of SR-GPT and four baselines on more than ten mainstream symbolic regression datasets. 
\label{tab1}}
\resizebox{14.9cm}{!}{
\begin{tabular}{cccccccc}
\toprule[1.45pt]
\toprule
Dataset & Dataset & Number & SR-GPT &  DGSR-MCTS & SPL  & TPSR & NeSymReS  \\ 
\toprule
Dataset-1 & Nguyen & 21 & $\textbf{96}_{\pm1.27} \%$ & $93_{\pm2.52}\%$ & $91_{\pm3.46}\%$ & $91_{\pm4.17}\%$& $56_{\pm3.67}\% $ \\
Dataset-2 & Keijzer & 15 & $\textbf{80}_{\pm2.41}\% $& $76_{\pm3.26} \%$& $67_{\pm3.24}\%$ & $77_{\pm6.33}\%$ & $52_{\pm5.93}\%$ \\
Dataset-3 & Korns & 15 & $\textbf{77}_{\pm1.99}\%$  & $ 71_{\pm2.42}\%$ & $ 69_{\pm4.10}\% $  & $ 66_{\pm5.21}\% $ & $ 27_{\pm5.29}\% $ \\
Dataset-4 & Constant & 8 & $\textbf{89}_{\pm3.44}\%$ & $83_{\pm3.23}\%$& $82_{\pm4.23}\% $& $84_{\pm5.14}\%$ & $22_{\pm5.17}\%$  \\
Dataset-5 & Livermore & 22 & $\textbf{91}_{\pm1.07}\%$& $86_{\pm2.13}\%$&$ 81_{\pm2.46}\%$& $83_{\pm2.17}\%$ & $28_{\pm3.58}\%$ \\
Dataset-6 & Vladislavleva & 8 & $\textbf{57}_{\pm2.90}\%$ &$ 51_{\pm2.28}\%$ & $39_{\pm3.29}\%$ & $42_{\pm2.47}\% $ & $13_{\pm3.20}\%$  \\
Dataset-7 & R & 6 & $\textbf{50}_{\pm1.82}\%$ & $45_{\pm1.81}\% $& $32_{\pm2.61}\% $& $40_{\pm1.33}\%$ & $4_{\pm3.18}\%$  \\
Dataset-8 & Jin & 6 & $\textbf{64}_{\pm1.18}\%$ & $52_{\pm1.42}\% $& $43_{\pm2.41}\%$ & $54_{\pm1.32}\%$& $11_{\pm3.01}\% $\\
Dataset-9 & Neat & 9 & $\textbf{71}_{\pm2.73}\%$ & $65_{\pm1.38}\% $& $65_{\pm2.47}\%$ & $63_{\pm1.34}\% $& $21_{\pm3.15}\% $\\
Dataset-10 & AIFeynman & 103 & $\textbf{65}_{\pm1.22}\% $& $57_{\pm2.57}\%$& $44_{\pm4.10}\%$ & $49_{\pm3.19}\% $&
$18_{\pm4.62}\% $\\
Dataset-11 & Others & 9 & $\textbf{84}_{\pm1.18}\%$ & $75_{\pm1.22}\%$& $74_{\pm2.19}\% $& $67_{\pm1.29}\%$ & $24_{\pm2.73} \%$\\

\cline{4-8} 
 & \multicolumn{2}{r}{Average} &$\textbf{74.90}_{\pm10.62} \%$ &$68.5_{\pm11.1}\%$ & $ 63.1_{\pm13.4}\%$ & $65.1_{\pm10.8}\%$ & $25.1_{\pm10.74}\% $
\end{tabular}

}
\end{table*}  
\begin{figure*} 
    \centering
    \vspace{-0.5cm}
    \hspace{-0.6cm}
    \setlength{\belowcaptionskip}{-0.2cm} 
	  \subfloat[]{       \hspace{-0.2cm}\includegraphics[width=0.326\linewidth]{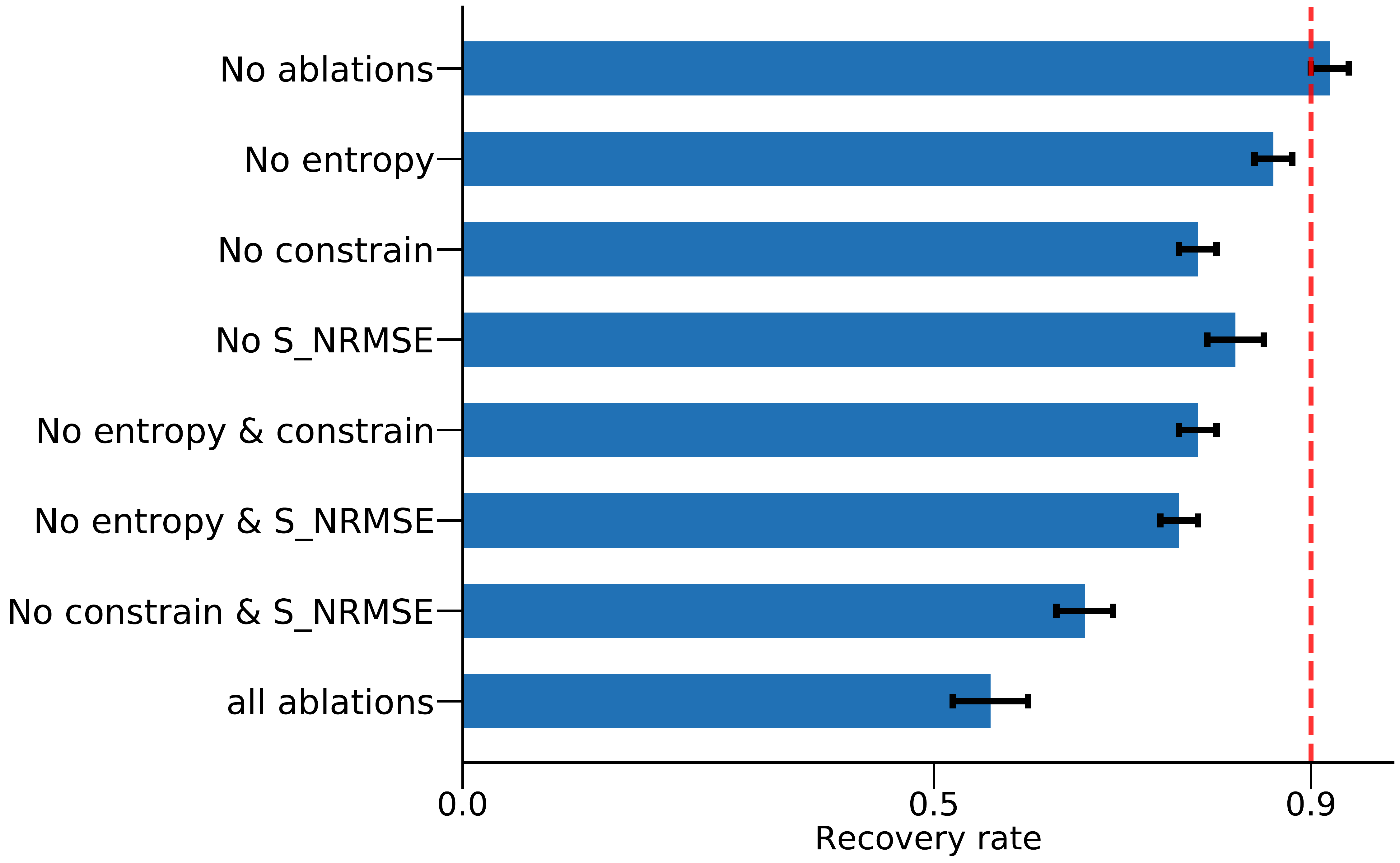} \label{fig3a}}
        \subfloat[]{
        \hspace{-0.05cm}\includegraphics[width=0.326\linewidth]{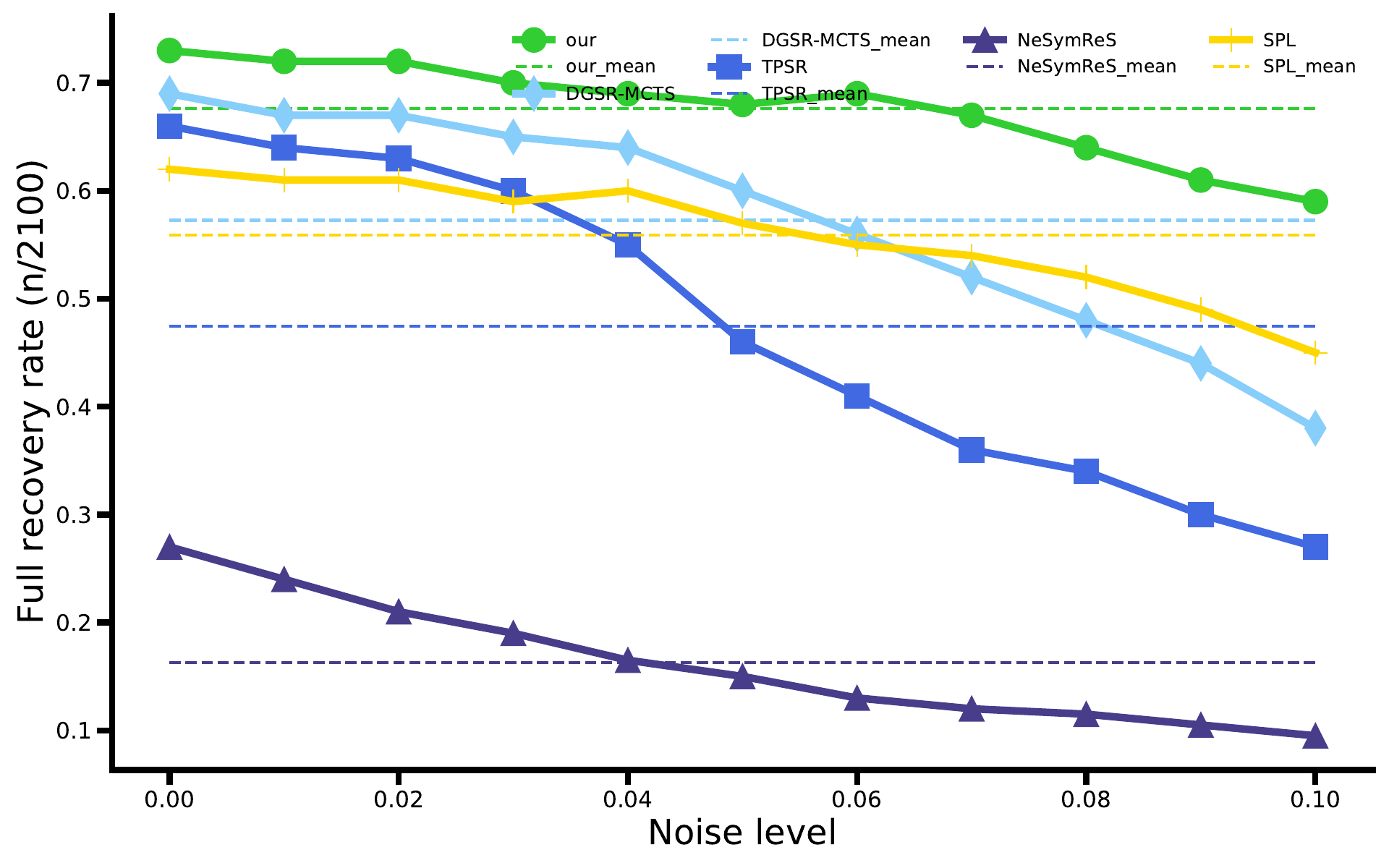}\label{fig3b}}   
        \subfloat[]{
    \hspace{-0.1cm}\includegraphics[width=0.400\linewidth]{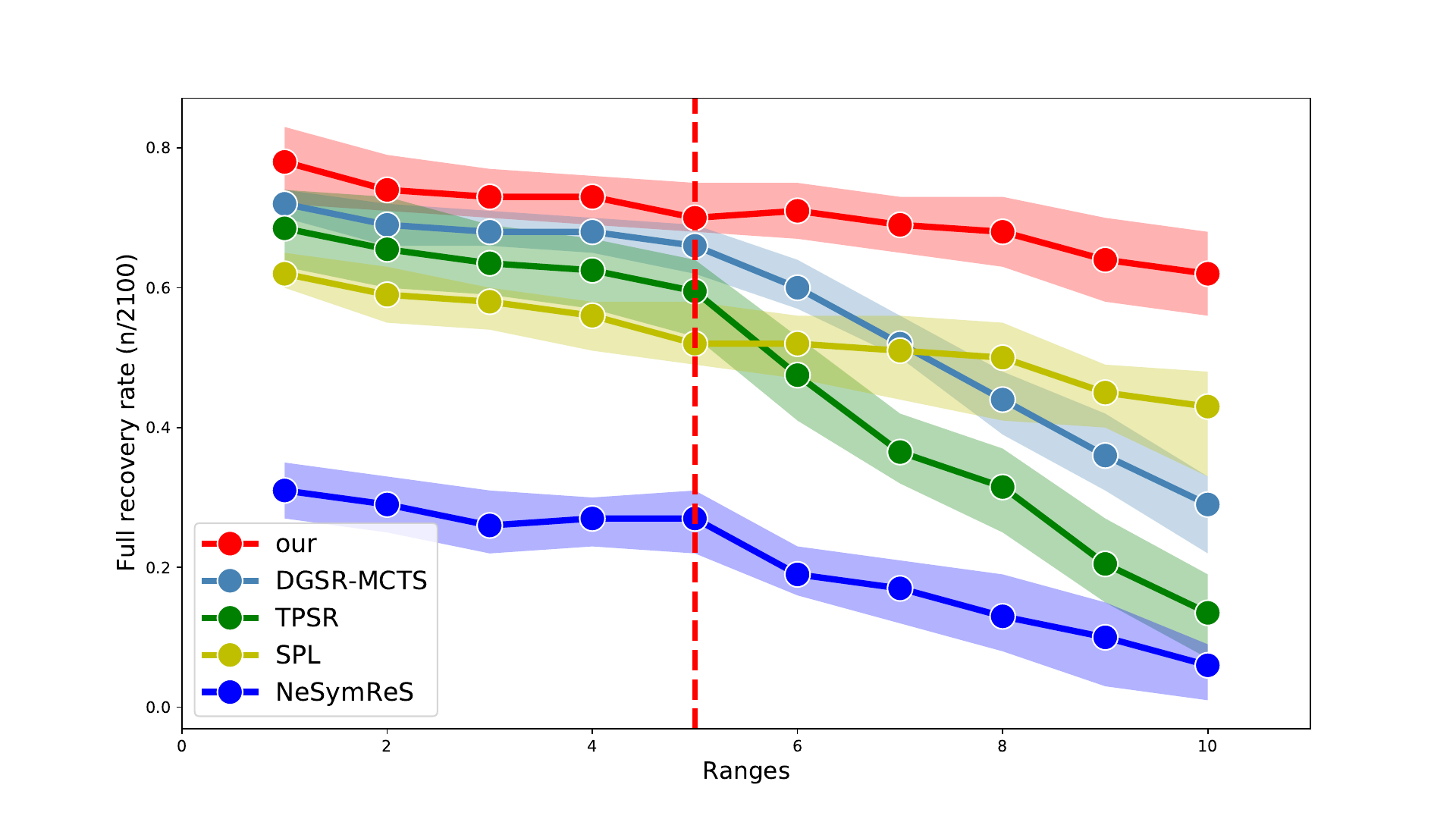}\label{fig3c}}   
	  \caption{
   Fig (a) illustrates the recovery for various ablations of SR-GPT on all Nguyen benchmarks, with error bars indicating the standard errors. Fig (b) describes the recovery rate of SR-GPT and four other excellent algorithms on all benchmarks under different levels of noise. Fig (c) It shows the general ability of five algorithms, from which we can see that SR-GPT is better than other algorithms in general ability. }

\label{fig3} 
\end{figure*}

\section{Results}
\textbf{Evaluating SR-GPT}. We evaluated the performance of SR-GPT on more than ten classic datasets in the field of symbolic regression. These datasets are labeled $Nguyen, Keijzer, Korns, Constant, Livermore, R\\ Vladislavlev, Jin$, $Neat$, $AI Feynman$, and $Others$. The datasets mentioned above collectively contain a total of 222 test expressions, The specific details are shown in table \cref{a-tab1,a-tab2,a-tab3} in the appendix.
We compare SR-GPT with four symbol regression algorithms that have demonstrated high performance:
\begin{itemize}
\item \textbf{SPL}. An excellent algorithm that successfully applies the traditional MCTS to the field of symbolic regression.
\item \textbf{DGSR-MCTS}. A symbolic regression algorithm inspired by GP trains a policy network to predict 'mutation points', which is then used to guide MCTS.
\item \textbf{NeSymReS}. This algorithm is categorized as a large-scale pre-training model.
\item \textbf{TPSR}. A symbolic regression algorithm that uses pre-trained models like NeSymReS to guide MCTS. 

\end{itemize}
In order to test the ability of each algorithm to fully recover expression using only local data, we sampled only 20 points for variable $X$ within the interval $[-1,1]$ and obtained the corresponding true value $y$. 
We use the strictest definition of recovery: exact symbolic equivalence, as determined using a computer algebra system, e.g. SymPy \citep{meurer2017sympy}.
We tested each expression 100 times using the aforementioned algorithms and recorded the number of times, denoted by $\mathcal{N}_{full}$, that expression was completely recovered (The expression symbols, constant positions, and constant values are all the same). Finally, we evaluated the performance of each algorithm by comparing the full recovery rate, calculated as $\mathcal{N}_{full}/100$. In Table \ref{tab1}, we present the recovery rates of each algorithm on all benchmark datasets. The performance of SR-GPT on recovery rate is slightly better than the other four advanced algorithms. In addition, the $R^2$ of SR-GPT on the AI Feynman dataset is presented in \ref{EqTableA}\ref{EqTableB}. As the training progresses, the algorithm's reward function fluctuation is illustrated in the line graph (convergence proof) as depicted in Figures \ref{fig-r} of the appendix. From the reward fluctuation line chart, one can discern that as training ensues, the reward values exhibit an oscillatory ascent, eventually culminating in a state of equilibrium. This corroborates the efficacy of our algorithm in guiding symbolic search, and it also validates that our algorithm ultimately attains convergence.
\\
\textbf{Ablation studies}.
SR-GPT includes a series of small but important components, in addition to its main algorithm. We developed a series of ablation experiments to verify the effect of each component on the algorithm performance. The image in Fig \ref{fig3a} shows the SR-GPT performance change on all Nguyen benchmarks after the different component combinations are removed. Where "No entropy" means that the information entropy loss is not applied to the loss. "No constrain" means no constraints are applied. "No S\_NRMSE" means that the $S_{NRMSE}$ loss function was not applied. As can be seen from the figure, although there is no catastrophic decline in algorithm performance after different components are removed, the recovery rate is significantly reduced compared with that without the ablation experiment.

\begin{table}[htbp]
\caption{This demonstrates the impact of incorporating the information entropy of predicted probability $p$ into the loss function during GPT training on the efficiency of the algorithm.
\label{tab2}}
{
\resizebox{8cm}{!}{
\begin{tabular}{ccccc}
\toprule[1.2pt]
Benchmark & \multicolumn{2}{c}{Mean entropy}& \multicolumn{2}{c}{Time(s)} \\
\cline{1-5}
   & Yes & No  & Yes & No \\
\cline{1-5}
Nguyen-1  & $\textbf{1.50}$ & $2.16$ & $\textbf{14.2}$   & $25.6$  \\
Nguyen-2  & $\textbf{1.62}$ & $2.43$ & $\textbf{115.34}$ & $168.16$\\
Nguyen-3  & $\textbf{1.82}$ & $2.96$ & $\textbf{268.42}$ & $484.76$ \\   
Nguyen-4  & $\textbf{2.19}$ & $3.02$ & $\textbf{268.42}$ & $484.76$ \\ 
\cline{2-5} 
 {Average} & $\textbf{1.78}$ &$2.64$ & $\textbf{132.65}$ &$226.17$  
\end{tabular}
}
}
\end{table}
\textbf{Anti-noise experiment}
In the natural world, most of the data we get has some level of noise. Therefore, the anti-noise ability is an important index to test whether an algorithm can solve real problems. We conducted anti-noise experiments on the SR-GPT algorithm and four other advanced algorithms on Nguyen benchmarks. The noisy data\citep{shang2018learning} were randomly sampled from the range of $[-level*scale, level*scale]$ where $level \in [0,0.1]$ represents the noise level, and the $scale$ is equal to $max(y)-min(y)$. At each noise level, we ran each expression 100 times. Then, we calculated the recovery rate of each algorithm on the expressions at each noise level. As shown in Fig \ref{fig3b}, SR-GPT exhibits outstanding performance in anti-noise ability. It outperforms all other algorithms at all noise levels except for slightly lower performance than DSO at noise levels of 0.01 and 0.02.\\
\textbf{Ablation experiment for information entropy} In order to increase the "confidence" of the GPT in selecting the correct symbol, it is highly preferable that the predicted probability distribution $p$ be concentrated on the correct symbol. We incorporated the information entropy of the probability distribution $p$, represented as $-p^T log(p)$, into the loss function. We performed ablation experiments on three expressions Nguyen-1 to Nguyen-3. During the experimentations, the predicted values $[p_1, p_2, p_3,\dots]$ were retained after every prediction of the GPT. The information entropy $E_i$ for each $p_i$ was then calculated as $E_i = -p_i^T \log p_i$. Finally, the mean information entropy, $E_{mean}$, was determined by averaging the information entropy for each $p_i$. Additionally, we monitored the impact of incorporating the information entropy of the predicted probability distribution $p$ into the loss function on the algorithm's efficiency. As shown in Table \ref{tab2}, incorporating the information entropy of $p$ into the loss function makes the GPT more ``confident'' in its predictions, resulting in a lower information entropy of the predicted probabilities $p$. This indicates that the GPT is more certain in selecting a symbol. Meanwhile, the GPT's increased "confidence" can result in improved algorithm efficiency. Under equal conditions, we are able to find the target expression more quickly. 
\\
\textbf{Running efficiency test}
In the field of symbol regression, running efficiency is also an important indicator to measure the performance of an algorithm. To test the efficiency of the five algorithms, we first filtered out the expressions in each dataset that each algorithm was able to fully recover. The time taken by each algorithm to recover these expressions is then counted. And then take the average. The detailed results are shown in Table 3. Through the table data, we can find that SR-GPT is slightly lower than TPSR, and DGSR-MCTS in efficiency. However, it is higher than SPL and better than all baselines in full recovery rate Table\ref{tab1}, generality, and noise immunity. It is proved that our algorithm truly achieves a balance between efficiency and generality while ensuring high performance.

\begin{table}[H]
  \centering  
  \caption{Running time (s) of SR-GPT and other baselines on Nguyen dataset}
    \begin{tabular}{lcccc}
    Benchmark & SR-GPT &DGSR-MCTS&SPL& TPSR\\
    \toprule
    Nguyen-1 & 14.22   & \textbf{9.05}    &18.29  &  11.55\\
    Nguyen-2 & 115.34  &\textbf{86.59}    &188.23 &  96.27\\
    Nguyen-3 & 132.64  &\textbf{108.88}   &316.35 &  128.20 \\
    Nguyen-4 &268.42   &\textbf{199.41}   &589.54 &  228.35\\
    Nguyen-5 & 624.45  &568.05   &831.28 &  \textbf{539.55}\\
    Nguyen-6 & 136.24  &\textbf{88.79}    &174.61 &  101.27\\
    Nguyen-7 & 36.24   &28.38    &84.99  &  \textbf{22.20}\\
    Nguyen-8 & \textbf{1.22}    &4.41     &1.23   &  13.35\\
    Nguyen-9 & 17.18   &16.05    &24.14  &  \textbf{14.66}\\
    Nguyen-10 & 38.28  &33.79    &53.24  &  \textbf{26.27}\\
    Nguyen-11 & 62.44  &\textbf{48.38}    &80.73  &  51.20\\
    Nguyen-12 & 323.42 & 278.35  &487.90 &  \textbf{277.63}\\
    \toprule
    \textbf{Average} & 147.50 &\textbf{122.51}&237.54&125.87\\
    \end{tabular}
  \label{tab:runtime}
\end{table}
\textbf{General capability test}
For symbolic regression algorithms, in addition to noise immunity. The ability to adapt to different intervals of data is also very important because the data in nature is very diverse, we call this ability the "universal ability" of the model. In order to test the general ability of the various models. We are in $[-1,1],... ,[-5,5]$ a total of 3 million training samples are sampled over the five intervals. It is used to train DGSR-MCTS, TPSR, and NeSymReS. Then the five algorithms are tested on the interval $[-1,1],...,[-10,10]$. The results are shown in Figure 3. We can see that SR-GPT and SPL are better than the other three algorithms in general ability. Because for the pre-trained model, its training data is only in $[-1,1],...,[-5,5]$. Therefore, it only performs well in these intervals, and cannot generalize to other intervals, and performs poorly beyond $[-5,5]$. Even in the range $[-1.5,1.5]$ it does not perform well.\\
\textbf{Average $R^2$ on Various Datasets}
To assess the goodness of fit of SR-GPT on the datasets, we also recorded the average $R^2$ of SR-GPT on various testing datasets. From the table, it can be observed that SR-GPT achieved an average R2 exceeding 0.99 on the majority of the datasets. This suggests that while SR-GPT may not be able to fully recover the original formula of certain expressions, it can still find an equivalent expression that fits the observed data well.  
{
\begin{table}[H]
  \centering  
  \caption{Average Coefficient of Determination ($R^2$) on Various Datasets}
    \begin{tabular}{lc}
    Benchmark & $R^2$ \\
    \toprule
    Nguyen & 0.9999 \\
    Keijzer & 0.9991 \\
    Korns & 0.9982 \\
    Constant & 0.9991 \\
    Livermore & 0.9998 \\
    Vladislavlev & 0.9831 \\
    R & 0.9702 \\
    Jin & 0.9888 \\
    Neat & 0.9763 \\
    AI Feynman & 0.9960 \\
    Others & 0.9982 \\
    \toprule
    \textbf{Average} & \textbf{0.9917} \\
    \end{tabular}
  \label{tab:runtime}
\end{table}}

\section{Disscusion}
We propose a new symbolic regression framework SR-GPT. The results are state-of-the-art on a series of benchmark symbolic regression tasks. In SR-GPT, we introduce an GPT to guide MCTS in searching for good expressions. The data obtained from the MCTS simulation is then used to train the GPT, which in turn provides better guidance for MCTS. To make the GPT more ``confident'', we introduce the information entropy loss of the predicted probability distribution $p$ in the loss function. Experimental results demonstrate that this operation can effectively improve the efficiency of the algorithm search. Secondly, in order to improve the problem of variable dropout that often occurs when the algorithm deals with multi-variable problems. We propose a new loss function, $S_{NRMSR}$, which not only considers $y$ and $\hat{y}$ but also considers $x_n$ using the Quasi-Euclidean distance between the real and predicted points in a multi-dimensional space as the loss. Our ablation experiments have demonstrated that introducing the proposed $S_{NRMSR}$ loss function yields a noteworthy enhancement in the algorithm's expression recovery rate, compared to using a regular loss function. Furthermore, this loss function effectively addresses the challenge of variable omission that often arises when dealing with multi-variable problems. \\
When SR-GPT deals with data with too many variables, the search efficiency is low because of the large search space, and sometimes the target expression can not be obtained. In order to solve this problem, next we intend to pre-train the policy network with a large amount of data to improve the performance and efficiency of the algorithm.\\
Our future research will focus on the following aspect. In the physical world, real physical quantities have specific units, for instance, the unit of velocity is measured in meters per second (m/s). Hence, the expressions that we discover should be a variable measured in meters divided by a variable measured in seconds. With this prior knowledge, we can significantly enhance search efficiency. Therefore, in the following research, we will consider incorporating physical constraints into the search process based on the laws of the real physical world. This will result in a more scientifically sound and efficient symbol regression algorithm.

\section{Acknowledgments}
This work was supported in part by the National Natural Science Foundation of China under Grant 92370117, in part by CAS Project for Young Scientists in Basic Research under Grant YSBR-090 and in part by the Key Research Program of the Chinese Academy of Sciences under Grant XDPB22

\nocite{langley00}

\bibliography{example_paper}
\bibliographystyle{icml2024}

\appendix
\newpage
\onecolumn
\appendix
\renewcommand\thesection{\Alph{section}} 
\counterwithin{figure}{section} 
\counterwithin{table}{section}
\section{Appendix: Pseudocode for the SR-GPT}
\textbf{Algorithm 1} presents an overview of the SR-GPT framework. Prior to selecting a symbol, the $UseConstraint(\tau, s, \pi)$ function is applied to enforce constraints. The $counter$ is updated after each symbol selection, and we check whether the $counter$ is zero. If the $counter$ is zero, the reward value of the obtained expression is computed and backpropagation is performed. If the current symbol is a leaf node but not a terminal node, we expand and evaluate the node.

\begin{algorithm}[H]
  \caption{SR-GPT}
  \label{algorithm1}
  \SetKwInput{KwData}{Data}
  \SetKwInput{KwResult}{Result}
  \KwData{$X = [x_1, x_2, \dots, x_n]$; $y = [y_1, y_2, \dots, y_n]$; $S = \{+, -, \times, \div, \dots\}$}
  \KwResult{Find an expression such that $y = f(X)$}
  
  initialization;\\
  \While{Reward $\neq 1$}{
    \Repeat{Find the target expression}{
      \textit{Self-Search:}\\
      UseConstraint($\tau, s, \pi$); \tcp{Algorithm UseConstraint is referred with \ref{algorithm3}}
      $symbol = \arg\max (\pi)$; \tcp{Choosing the symbol with the highest probability}
      $\tau.\text{append}(symbol)$;\\
      $counter = counter + \text{Arity}(symbol) - 1$; \tcp{Whether or not the expression is complete}
      \If{$counter = 0$}{
        $z = \frac{1}{1 + S_{\text{NRMSE}}}$; \tcp{Calculating rewards}
        \If{$z > T$}{
          break;  \tcp{Terminate the program upon achieving expected rewards}
        }
        \textbf{Backpropagate:} $z$; \tcp{Backpropagate the final reward}
        Storing data: $[s, \pi, z]$;\\
        \textit{Train Neural Network:} $\mathcal{N}_{\theta} \rightarrow \mathcal{N}_{\theta \text{NEW}}$; \tcp{Further training of GPT}
      }
      \textit{MCTS:}\\
      \textbf{Expand and evaluate:}\\
      $parent \, \|\, sibling = \text{ParentSibling}(\tau_t)$; \tcp{Get the neural network input}
      $(p, v) = \mathcal{N}_{\theta}(parent \,\|\, sibling)$; \tcp{Calculating probability distribution p and evaluat value v with GPT}
      \For{$j \leftarrow 2$ \KwTo $n_{\text{evaluate}}$}{
        \eIf{$\text{current\_node} = \text{leaf\_node}$ \textbf{and} $counter \neq 0$}{
          \textbf{Expend:} $p$; \tcp{Expanding leaf nodes with probability p}
        }{
          \textbf{Select:}\\
          $a_{t+1} = \arg\max(\mathcal{UCT}(s_t, a_t))$; \tcp{Selecting the symbol with the largest UCT value as the next symbol}
        }
        \textbf{Backpropagate:} $v$; \tcp{Backpropagate the evaluate value v}
      }
    }
  }
\end{algorithm}

\textbf{Algorithm 2} describes the function $ParentSibling(\tau) $ used in Algorithm 1 to find the parent and sibling nodes of the next symbol to the sample. This algorithm uses the following logic: If the last symbol in the partial traversal is a unary or binary operator and has no siblings, it is the parent node and its sibling is set to empty. Otherwise, the algorithm iterates forward in the traversal until it finds a node with unselected children. This node is the parent and the subsequent node is the sibling.\\
    

\begin{algorithm}[H]
\caption{ParentSibling($\tau$) (To retrieve the father and sibling nodes as inputs for a GPT)}
\label{algorithm2}
  \SetKwInput{KwInput}{Input}
  \SetKwInput{KwOutput}{Output}
  \KwInput{Partially sampled traversal $\tau$}
  \KwOutput{Concatenated parent and sibling nodes of the next nodes to be generated}
  
  $T \leftarrow \text{len}(\tau)$ \tcp*{Length of partial traversal}
  $counter \leftarrow 0$ \tcp*{Initializes a counter with no selected number of nodes}
  
  \If{$\text{Arity}(\tau[T])>0$}{
      $parent \leftarrow \tau[T]$ \;
      $sibling \leftarrow \text{empty}$ \;
  }
  
  \For{$i\leftarrow T$ \KwTo 1}{
    $counter \leftarrow counter + \text{Arity}(\tau[i])-1$ \tcp*{Update counter of unselected nodes}
    \If{$counter = 0$}{
      $parent \leftarrow \tau[i]$ \;
      $sibling \leftarrow \tau[i+1]$ \;
    }
  }
\end{algorithm}
\textbf{Algorithm 3} demonstrates the application of a series of constraints during the symbol generation process. The specific steps are as follows: we first obtain the types of symbols in our symbol library, and then based on the current state, we sequentially determine whether each function in the symbol library should be "restricted". If a symbol is restricted, we set the probability of selecting that symbol to zero and finally normalize the probabilities of all symbols.\\

\begin{algorithm}[H]
\label{algorithm3}
  \SetKwInput{KwInput}{Input}
  \SetKwInput{KwOutput}{Output}
  \KwInput{The simulated probability $\pi$; partially sampled traversal $\tau$; Used symbol library $S$}
  \KwOutput{The probability distribution $\pi$ adjusted according to the constraints}
  
  $L \leftarrow len(S)$ \tcp*{Length of $S$}
  
  \For{$i\leftarrow 1$ \KwTo $L$}{
    \If{Constraint($\tau,S[i]$)}{
      $\pi[i] \leftarrow 0$ \tcp*{Sets the restricted symbol probability to zero}
    }
  }
  $\pi \leftarrow \frac{\pi}{\sum_{i=1}^{L}\pi[i]}$ \tcp*{The probability is normalized}

  \caption{UseConstraints($\tau, S, \pi$)}
\end{algorithm}

 \textbf{Algorithm 4} describes the Arity(s) function used in Algorithm 1, which obtains the arity of an operator. Specifically, for a variable $[x_1]$ and a constant $[c]$, Arity(s)=0. For a unary operator $[sin, cos, exp, ln, sqrt]$, Arity(s)=1. Similarly, for a binary operator, Arity(s)=2, and so on for other operators.\\

  
  

\begin{algorithm}[H]
  \SetAlgoLined
  \DontPrintSemicolon
  \SetKwInput{Input}{Input}
  \SetKwInput{Output}{Output}
  \SetKwFunction{FSelect}{select}
  \SetKw{KwIn}{in}
  \SetKwProg{Fn}{Function}{:}{}
  \SetKwIF{If}{ElseIf}{Else}{if}{:}{else if}{else}{endif}

  \Input{Newly selected operator symbol $s$}
  \Output{The arity of an operator}
  
  \Fn{Arity}{$s$}{
  
  $s \leftarrow$ \FSelect{$S$}\;
  \tcc{Selecting new symbols}

  \If{$s \KwIn \{x_1,x_2,...,x_n,c\}$}{
    \KwRet 0\tcp*{If the symbol is a variable or a constant, the arity is 0}
  }
  
  \If{$s \KwIn \{sin, cos, exp, log, sqrt\}$}{
    \KwRet 1\tcp*{If the operator is unary, the arity is 1}
  }
  
  \If{$s \KwIn \{+,-,*,/\}$}{
    \KwRet 2\tcp*{If the operator is a binary operator, it returns 2}
  }
  
  }
  \caption{Compute the arity of an operator $s$}
  
\end{algorithm}

\newpage
\section{Appendix: Reward variation curve (On Dataset Nguyen).} 
\begin{figure*}[h]
    \centering
	  \subfloat[Nguyen-1]{
      \includegraphics[width=0.26\linewidth]{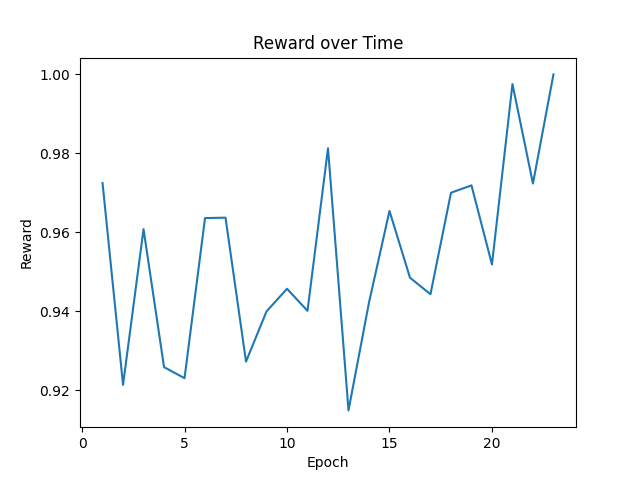}}
    \label{1a}
	  \subfloat[Nguyen-2]{
        \includegraphics[width=0.26\linewidth]{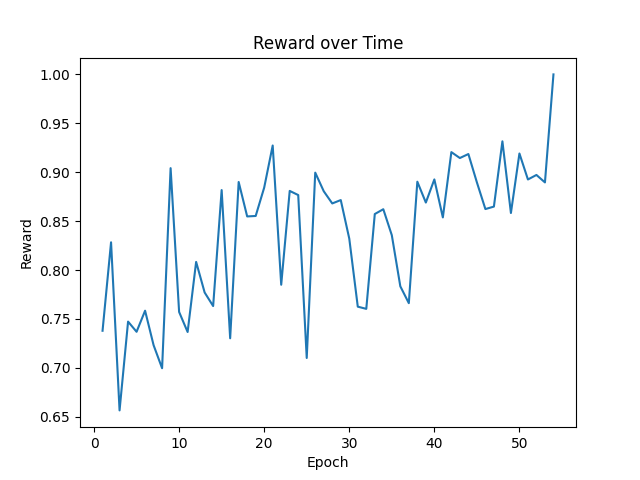}}
    \label{1b}
    \subfloat[Nguyen-3]{
        \includegraphics[width=0.26\linewidth]{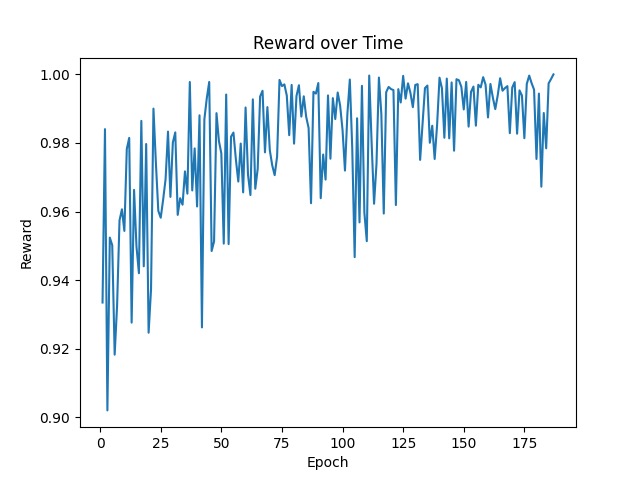}}
    \label{1b}\\
    \subfloat[Nguyen-4]{
        \includegraphics[width=0.26\linewidth]{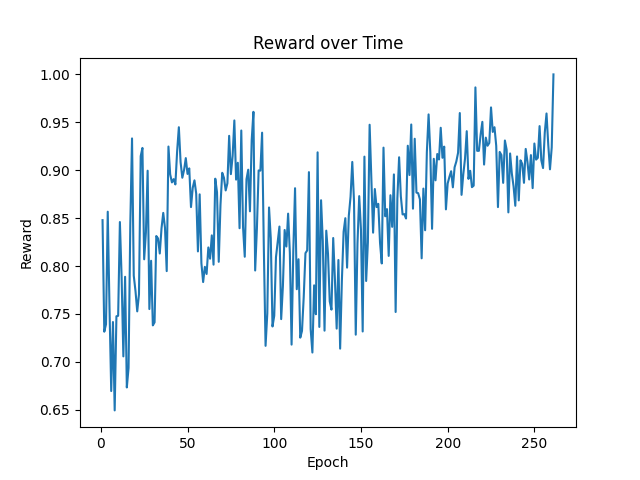}}
    \label{1b}
    \subfloat[Nguyen-5]{
      \includegraphics[width=0.26\linewidth]{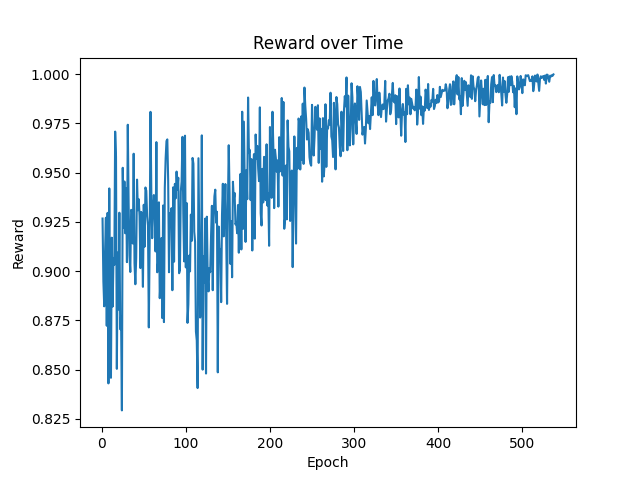}}
    \label{1a}
	  \subfloat[Nguyen-6]{
        \includegraphics[width=0.26\linewidth]{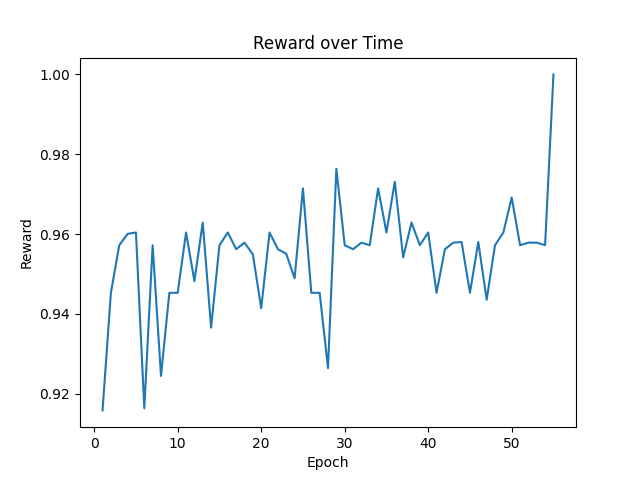}}
    \label{1b}\\
    \subfloat[Nguyen-7]{
        \includegraphics[width=0.26\linewidth]{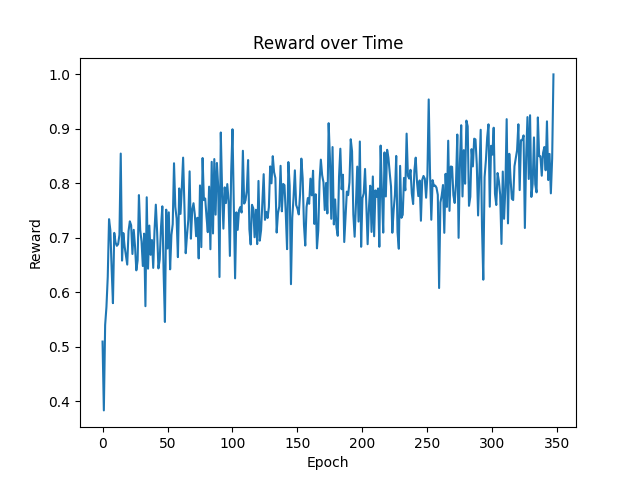}}
    \label{1b}
    \subfloat[Nguyen-8]{
        \includegraphics[width=0.26\linewidth]{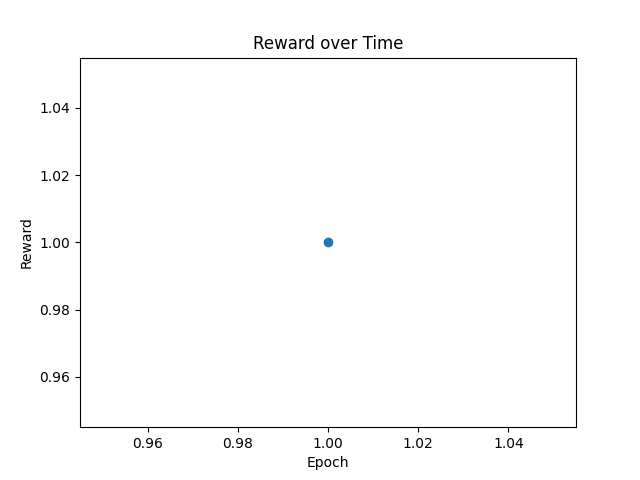}}
    \label{1b}
    \subfloat[Nguyen-9]{
        \includegraphics[width=0.26\linewidth]{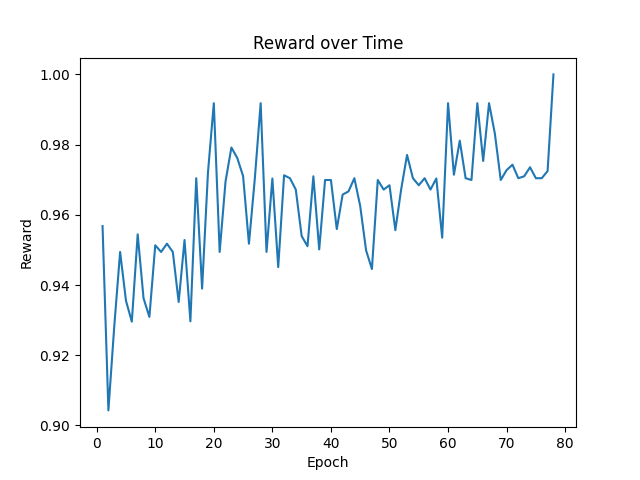}}
    \label{1b}\\
    \subfloat[Nguyen-10]{
        \includegraphics[width=0.26\linewidth]{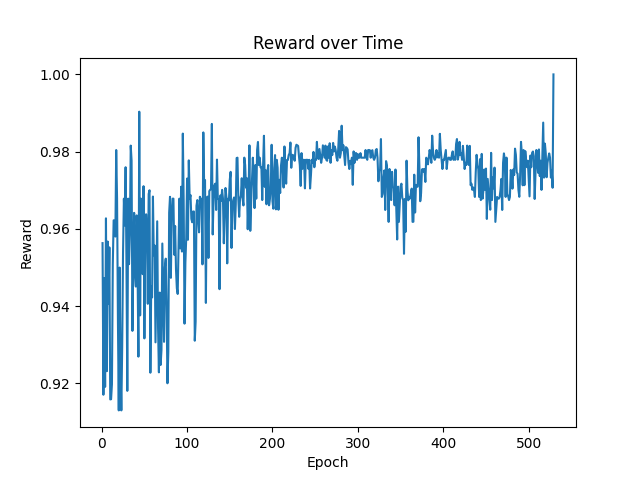}}
    \label{1b}
    \subfloat[Nguyen-11]{
        \includegraphics[width=0.26\linewidth]{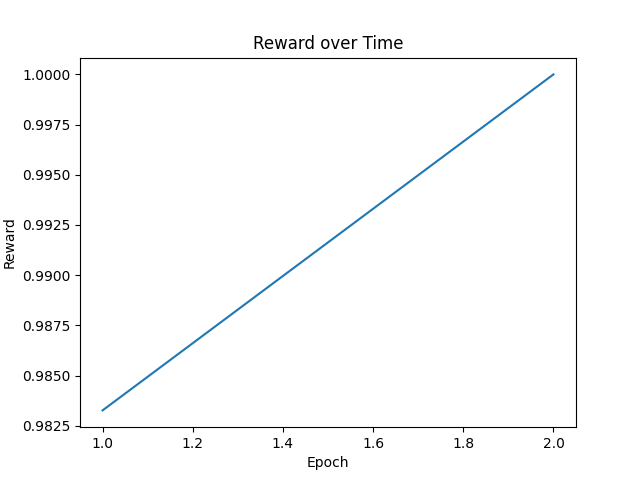}}
    \label{1b}
    \subfloat[Nguyen-12]{
        \includegraphics[width=0.25\linewidth]{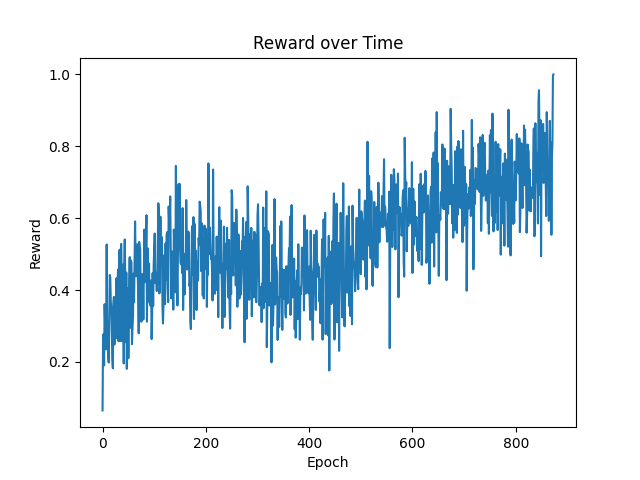}}
    \label{1b}\\
	  \caption{The series of figures above presents line graphs depicting the reward values of SR-GPT on the Nguyen dataset over time. As observed from the figures, throughout the search process, the reward values for all expressions demonstrate an oscillatory ascent with the increase in training iterations. Notably, Expression 8 is an exception due to its comparatively simple structure, achieving its best result within just one epoch.}
	  \label{fig-r} 
\end{figure*}
\newpage
\section{The resulting analyzability test}
\begin{figure*}[h]
    \centering
    \setlength{\belowcaptionskip}{-0.2cm} 
	  \subfloat[]{       \includegraphics[width=0.326\linewidth]{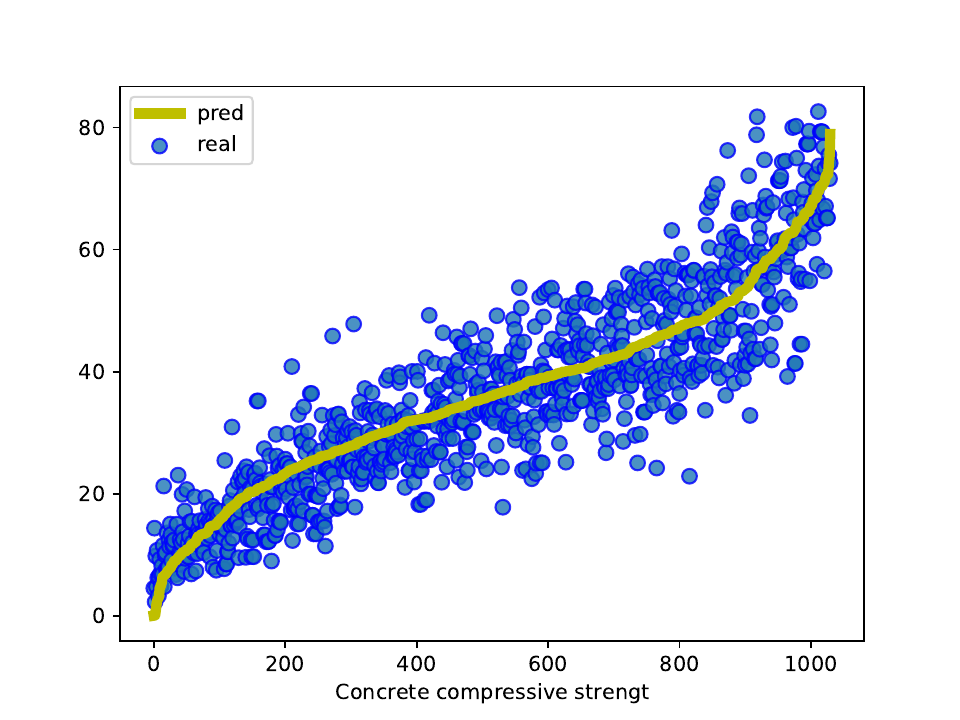} \label{fig3c}}
        \subfloat[]{
    \includegraphics[width=0.326\linewidth]{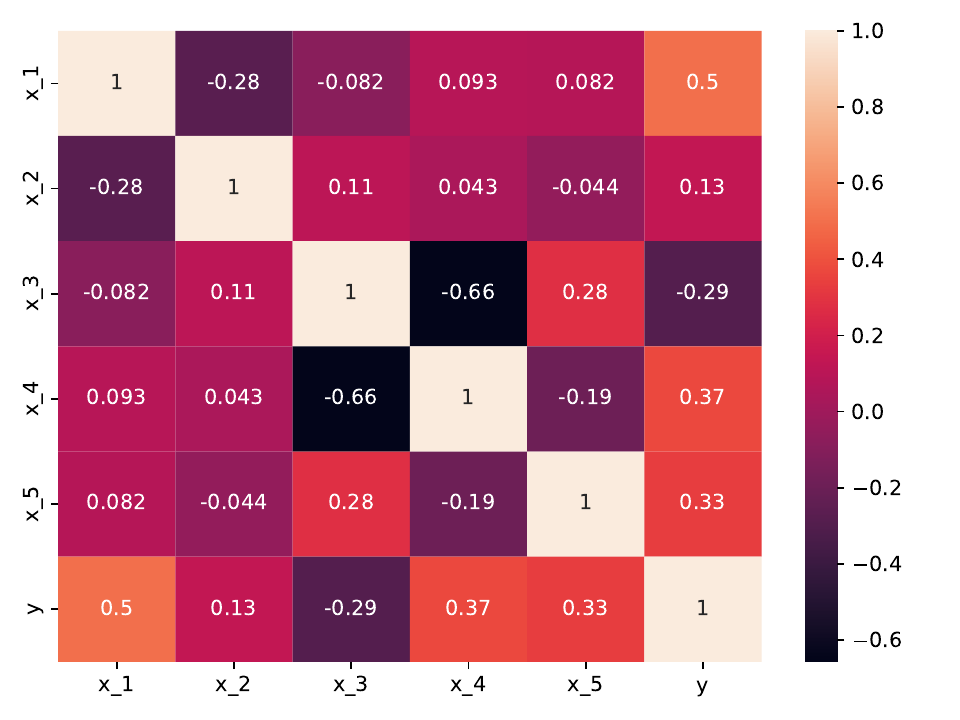}\label{fig_ham}}   
	  \caption{
   Fig (a) illustrates the recovery for various ablations of SR-GPT on all Nguyen benchmarks, with error bars indicating the standard errors. Fig (b) describes the recovery rate of SR-GPT and four other excellent algorithms on all benchmarks under different levels of noise. Fig (c) It shows the general ability of five algorithms, from which we can see that SR-GPT is better than other algorithms in general ability. }
\label{fig3} 
\end{figure*}

This Concrete compressive strength data set \citep{concrete} contains 5 quantitative input variables and 1 quantitative output variable. The 5 input variables are Cement ($x_1$), Blast Furnace Slag ($x_2$), Water ($x_3$), Superplasticizer ($x_4$), and Age ($x_5$). And the output variable is Concrete compressive strength ($y$). We use the aforementioned 5 input variables as input into SR-GPT to fit the output variable Concrete compressive strength. The final expression derived by SR-GPT is as shown in equation \ref{e7}.
\begin{equation}
\label{e7}
\displaystyle y = 0.66  x_4 + \frac{0.66(x_1 + x_2 -x_3 + 334)}{7.17 + \frac{17746.56}{x1*x5}}
\end{equation}
We can analyze the positive or negative correlation between the variables $[x1, x2,...,x_5]$ and $y$ very clearly from these equations. 
According to Formula 8, we can take the partial derivative of $y$ with respect to $x$, where $\frac{\partial y}{\partial x_1} = \frac{0.66(7.17 + \frac{17746.56}{x_1*x_5}) + \frac{17746.65}{x_1^2x_5}(0.66(x_1 + x_2 -x_3 + 334))}{{7.17 + \frac{17746.56}{x1*x5}}^2}>0$, $\frac{\partial y}{\partial x_2} = \frac{0.66}{7.17 + \frac{17746.56}{x_1*x5}}>0$, $\frac{\partial y}{\partial x_3} = \frac{-0.66}{7.17 + \frac{17746.56}{x_1x_5}}<0$, $\frac{\partial y}{\partial x_4} = 0.66>0$, $\frac{\partial y}{\partial x_5} = \frac{0.66(x_1 + x_2 -x_3 + 334)}{{7.17 + \frac{17746.56}{x_1x_5}}^2} \frac{17746.56}{ x_1^2x_5}>0$ Based on the information, we can conclude that $x_1,x_2,x_4,x_5$ is directly proportional to $y$,and  $x_3$ is inversely proportional to $y$. This is consistent with the correlation coefficient matrix\ref{fig_ham}, which indicates that the expression obtained by our algorithm is an analyzable and reliable expression. Figure \ref{fig3c} displays the fitted curve of Equation \ref{e7} to the data.

\newpage
\section{Appendix: Experimental details for each expression}
Table $B.1 - B.3$ shows some specific details of different expressions when tested.
The benchmark problem specifications for symbolic regression are as follows:
\begin{itemize}
\item Input variables are represented as [$x_1,x_2,...,x_n$]
\item $U(a,b,c)$signifies $c$ random points uniformly sampled between $a$ and $b$ for each input variable. Different random seeds are used for training and testing datasets.
\item $E(a,b,c)$ indicates $c$ points evenly spaced between $a$ and $b$ for each input variable. 
\item To simplify the notation, libraries are defined relative to a base library $[+,-,\times,\div, sin, cos, ln, exp, sqrt, x_1]$.
\item Any unassigned operand is represented by $\placeholder$, for instance, $\placeholder^2$ corresponds to the square operator.

\end{itemize}
\begin{table*}[htbp]
\centering
\caption{ Symbol library and value range of the three data sets Nguyen, Korns, and Jin. 
}
\begin{scriptsize}
\begin{tabular}{ccccc}
\toprule[1.45pt]
\toprule
Name & Expression & Dataset & Library \\ \hline
Nguyen-1 & $x_1^3+x_1^2+x_1$ & U$(-1, 1, 20)$ & \NguyenX \\
Nguyen-2 & $x_1^4+x_1^3+x_1^2+x_1$ & U$(-1, 1, 20)$ & \NguyenX \\
Nguyen-3 & $x_1^5+x_1^4+x_1^3+x_1^2+x_1$ & U$(-1, 1, 20)$ & \NguyenX \\
Nguyen-4 & $x_1^6+x_1^5+x_1^4+x_1^3+x_1^2+x_1$ & U$(-1, 1, 20)$ & \NguyenX \\
Nguyen-5 & $\sin(x_1^2)\cos(x)-1$ & U$(-1, 1, 20)$ & \NguyenX \\
Nguyen-6 & $\sin(x_1)+\sin(x_1+x_1^2)$ & U$(-1, 1, 20)$ & \NguyenX \\
Nguyen-7 & $\log(x_1+1)+\log(x_1^2+1)$ & U$(0, 2, 20)$ & \NguyenX \\
Nguyen-8 & $\sqrt{x}$ & U$(0, 4, 20)$ & \NguyenX \\
Nguyen-9 & $\sin(x)+\sin(x_2^2)$ & U$(0, 1, 20)$ & \NguyenXY \\
Nguyen-10 & $2\sin(x)\cos(x_2)$ & U$(0, 1, 20)$ & \NguyenXY \\
Nguyen-11 & $x_1^{x_2}$ & U$(0, 1, 20)$ & \NguyenXYCZ \\
Nguyen-12 & $x_1^4-x_1^3+\frac{1}{2}x_2^2-x_2$ & U$(0, 1, 20)$ & \Nn \\
\toprule
Nguyen-2$'$ & $4x_1^4+3x_1^3+2x_1^2+x$ & U$(-1, 1, 20)$ & \NguyenX \\
Nguyen-5$'$ & $\sin(x_1^2)\cos(x)-2$ & U$(-1, 1, 20)$ & \NguyenX \\
Nguyen-8$'$ & $\sqrt[3]{x}$ & U$(0, 4, 20)$ & \NguyenXCZ \\
Nguyen-8$''$ & $\sqrt[3]{x_1^2}$ & U$(0, 4, 20)$ & \NguyenXCZ \\
\toprule
Nguyen-1\textsuperscript{c} & $3.39x_1^3+2.12x_1^2+1.78x$ & U$(-1, 1, 20)$ & \NguyenXC \\
Nguyen-5\textsuperscript{c} & $\sin(x_1^2)\cos(x)-0.75$ & U$(-1, 1, 20)$ & \NguyenXC \\
Nguyen-7\textsuperscript{c} & $\log(x+1.4)+\log(x_1^2+1.3)$ & U$(0, 2, 20)$ & \NguyenXC \\
Nguyen-8\textsuperscript{c} & $\sqrt{1.23 x}$ & U$(0, 4, 20)$ & \NguyenXC \\
Nguyen-10\textsuperscript{c} & $\sin(1.5x)\cos(0.5x_2)$ & U$(0, 1, 20)$ & \NguyenXYC \\
\toprule
Korns-1 & $1.57+24.3*x_1^4$ & U$(-1, 1, 20)$ & \NguyenX \\
Korns-2 & $0.23+14.2\frac{(x_4+x_1)}{(3x_2)}$ & U$(-1, 1, 20)$ & \NguyenXYC \\
Korns-3 & $4.9\frac{(x_2-x_1+\frac{x_1}{x_3}}{(3x_3))}-5.41$ & U$(-1, 1, 20)$& \NguyenXYZ \\
Korns-4 & $0.13sin(x_1)-2.3$ & U$(-1, 1, 20)$ & \NguyenX \\
Korns-5 & $3+2.13log(|x_5|)$ & U$(-1, 1, 20)$ & \NguyenXC \\
Korns-6 & $1.3+0.13\sqrt{|x_1|}$ & U$(-1, 1, 20)$ & \NguyenXC \\
Korns-7 & $2.1(1-e^{-0.55x_1})$ & U$(-1,1 , 20)$ & \NguyenXC \\
Korns-8 & $6.87+11\sqrt{|7.23 x_1 x_4 x_5|}$ & U$(-1, 1, 20)$ & \NguyenXC \\
Korns-9 & $12\sqrt{|4.2x_1x_2x_2|}$ & U$(-1, 1, 20)$ & \NguyenXYC \\
Korns-10 & $0.81+24.3\frac{2x_{1}+3x_2^2}{4x_3^3+5x_4^4}$ & U$(-1, 1, 20)$ & \NguyenXYZDC \\
Korns-11 & $6.87+11cos(7.23x_1^3)$ & U$(-1, 1, 20)$ & \NguyenXYC \\
Korns-12 & $2-2.1cos(9.8x_1^3)sin(1.3x_5)$ & U$(-1, 1, 20)$ & \NguyenXYC \\ 
Korns-13 & $32.0-3.0\frac{tan(x_1)}{tan(x_2)}\frac{tan(x_3)}{tan(x_4)}$ & U$(-1, 1, 20)$ & \NguyenXYtantanhZD \\
Korns-14 & $22.0-(4.2cos(x_1)-tan(x_2))\frac{tanh(x_3)}{sin(x_4)}$ & U$(-1, 1, 20)$ & \NguyenXYtantanhZD \\
Korns-15 & $12.0-\frac{6.0tan(x_1)}{e^{x_2}}(log(x_3)-tan(x_4))))$ & U$(-1, 1, 20)$ & \NguyenXYtanZD \\ 
\toprule
Jin-1 & $2.5 x_1^4-1.3 x_1^3 +0.5 x_2^2 - 1.7x_2$ & U $(-3, 3, 100)$ & \Jintf \\
Jin-2 & $8.0 x_1^2 + 8.0 x_2^3 - 15.0$ & U $(-3, 3, 100)$ & \Jin \\
Jin-3 & $0.2 x_{1}^{3} + 0.5 x_{2}^{3} - 1.2 x_2 - 0.5 x_{1}$ & U$(-3, 3, 100)$ & \Jint \\

Jin-4 & $1.5 \exp{x} + 5.0 cos(x_2)$ & U$(-3, 3, 100)$ & \Jin \\

Jin-5 & $6.0 sin(x_1) cos(x_2)$ & U$(-3, 3, 100)$ & \Jin \\

Jin-6 & $1.35 x_1 x_2 + 5.5 sin((x_1 - 1.0)(x_2 - 1.0))$ & U$(-3, 3, 100)$ & \Jin \\   

\newline
\end{tabular}
\end{scriptsize}
\label{a-tab1}
\end{table*}

\begin{table*}[htpb]
\centering
\caption{
Symbol library and value range of the three data sets neat, Keijzer and Livermore.
}
\begin{scriptsize}
\begin{tabular}{ccccc}
\toprule[1.45pt]
\toprule
Name & Expression & Dataset & Library \\
\hline
Neat-1 & $x_1^4+x_1^3+x_1^2+x$ & U$(-1, 1, 20)$ & \NguyenX \\
Neat-2 & $x_1^5+x_1^4+x_1^3+x_1^2+x$ & U$(-1, 1, 20)$ & \NguyenX \\
Neat-3 & $\sin(x_1^2)\cos(x)-1$ & U$(-1, 1, 20)$ & \NguyenXOne \\
Neat-4 & $\log(x+1)+\log(x_1^2+1)$ & U$(0, 2, 20)$ & \NguyenXOne \\
Neat-5 & $2\sin(x)\cos(x_2)$ & U$(-1, 1, 100)$ & \NguyenXY \\
Neat-6 & $\sum_{k=1}^x \frac{1}{k} $ & E$(1, 50, 50)$ & \Keijzer \\
Neat-7 & $2 - 2.1\cos(9.8x_1)\sin(1.3x_2)$ & E$(-50, 50, 10^5)$ & \NguyenXY \\
Neat-8 & $\frac{e^{-(x_1)^2}}{1.2 + (x_2-2.5)^2}$ & U$(0.3, 4, 100)$ & \VladislavlevaB \\
Neat-9 & $\frac{1}{1+x_1^{-4}} + \frac{1}{1+x_2^{-4}}$ & E$(-5, 5, 21)$ & \NguyenXY \\
\toprule
Keijzer-1 & $0.3x_1sin(2\pi x_1)$ & U$(-1, 1, 20)$ & \NguyenXC \\
Keijzer-2 & $2.0x_1sin(0.5\pi x_1)$ & U$(-1, 1, 20)$ & \NguyenXC \\
Keijzer-3 & $0.92x_1sin(2.41\pi x_1)$ & U$(-1, 1, 20)$& \NguyenXC \\
Keijzer-4 & $x_1^3e^{-x_1}cos(x_1)sin(x_1)sin(x_1)^{2}cos(x_1)-1$ & U$(-1, 1, 20)$ & \NguyenXOne \\
Keijzer-5 & $3+2.13log(|x_5|)$ & U$(-1, 1, 20)$ & \NguyenXC \\

Keijzer-6 & $\frac{x1(x1+1)}{2}$& U$(-1, 1, 20)$& \NguyenXC \\
Keijzer-7 & $log(x_1)$ & U$(0,1 , 20)$ & \NguyenX \\
Keijzer-8 & $\sqrt{(x_1)}$ & U$(0, 1, 20)$ & \NguyenX \\
Keijzer-9 & $log(x_1+\sqrt{x_1^2}+1)$ & U$(-1, 1, 20)$ & \NguyenXY \\
Keijzer-10 & $x_{1}^{x_2}$ & U$(-1, 1, 20)$ & \NguyenXYCZ \\
Keijzer-11 & $x_1x_2+sin((x_1-1)(x_2-1))$ & U$(-1, 1, 20)$ & \NguyenXYOne \\
Keijzer-12 & $x_1^4-x_1^3+\frac{x_2^2}{2}-x_2$ & U$(-1, 1, 20)$ & \Jintf \\ 
Keijzer-13 & $6sin(x_1)cos(x_2)$ & U$(-1, 1, 20)$ & \NguyenXYC \\
Keijzer-14 & $\frac{8}{2+x_1^2 + x_2^2}$ & U$(-1, 1, 20)$ & \NguyenXYC \\
Keijzer-15 & $\frac{x_1^3}{5}+\frac{x_2^3}{2}-x_2-x_1$ & U$(-1, 1, 20)$ & \Jint \\ 

\toprule
Livermore-1 & $\frac{1}{3}+x_1+sin(x_1^2))$ & U$(-3, 3, 100)$ & \nolog \\
Livermore-2 & $sin(x_1^2)*cos(x1)-2$ & U$(-3, 3, 100)$ & \nolog \\
Livermore-3 & $sin(x_1^3)*cos(x_1^2))-1$ & U$(-3, 3, 100)$ & \nolog \\
Livermore-4 & $log(x_1+1)+log(x_1^2+1)+log(x_1)$ & U$(-3, 3, 100)$ & \NguyenXOne \\ 
Livermore-5 & $x_1^4-x_1^3+x_2^2-x_2$ & U$(-3, 3, 100)$ & \Jintf \\
Livermore-6 & $4x_1^4+3x_1^3+2x_1^2+x_1$ & U$(-3, 3, 100)$ & \Jintf \\ 
Livermore-7 & $\frac{(exp(x1)-exp(-x_1)}{2})$ & U$(-1, 1, 100)$ & \NguyenXC \\ 
Livermore-8 & $\frac{(exp(x1)+exp(-x1)}{3}$ & U$(-3, 3, 100)$ &\NguyenXC \\
Livermore-9 & $x_1^9+x_1^8+x_1^7+x_1^6+x_1^5+x_1^4+x_1^3+x_1^2+x_1$ & U$(-1, 1, 100)$ & \NguyenX \\
Livermore-10 & $6*sin(x_1)cos(x_2)$ & U$(-3, 3, 100)$ & \NguyenXYC \\
Livermore-11 & $\frac{x_1^2 x_2^2}{(x_1+x_2)}$ & U$(-3, 3, 100)$ & \NguyenXY \\
Livermore-12 & $\frac{x_1^5}{x_2^3}$ & U$(-3, 3, 100)$ & \NguyenXY \\
Livermore-13 & $x_1^{\frac{1}{3}}$ & U$(-3, 3, 100)$ & \NguyenXCZ \\
Livermore-14 & $x_1^3+x_1^2+x_1+sin(x_1)+sin(x_2^2)$ & U$(-1, 1, 100)$ & \Jintf \\ 
Livermore-15 & $x_1^\frac{1}{5}$ & U$(-3, 3, 100)$ &\NguyenXCZ \\
Livermore-16 & $x_1^{\frac{2}{3}}$ & U$(-3, 3, 100)$ & \NguyenXCZ \\  
Livermore-17 & $4sin(x_1)cos(x_2)$ & U$(-3, 3, 100)$ & \NguyenXYC \\
Livermore-18 & $sin(x_1^2)*cos(x_1)-5$ & U$(-3, 3, 100)$ & \NguyenXC \\
Livermore-19 & $x_1^5+x_1^4+x_1^2 + x_1$ & U$(-3, 3, 100)$ & \NguyenX \\
Livermore-20 & $e^{(-x_1^2)}$ & U$(-3, 3, 100)$ & \NguyenXFOne \\
Livermore-21 & $x_1^8+x_1^7+x_1^6+x_1^5+x_1^4+x_1^3+x_1^2+x_1$& U$(-1, 1, 20)$ & \NguyenX \\
Livermore-22 & $e^{(-0.5x_1^2)}$ & U$(-3, 3, 100)$ &\NguyenXC \\
\newline
\end{tabular}
\end{scriptsize}
\label{a-tab2}
\end{table*}

\begin{table*}[htpb]
\centering
\caption{
Symbol library and value range of the three data sets Vladislavleva and others. }
\begin{scriptsize}
\begin{tabular}{ccccc}
\toprule[1.45pt]
\toprule
Name & Expression & Dataset & Library \\
\toprule
Vladislavleva-1 & $\frac{(e^{-(x1-1)^2})}{(1.2+(x2-2.5)^2))}$ & U$(-1, 1, 20)$ & \NguyenXC \\
Vladislavleva-2 & $e^{-x_1}x_1^3cos(x_1)sin(x_1)(cos(x_1)sin(x_1)^2-1)$ & U$(-1, 1, 20)$ & \Jin \\

Vladislavleva-3 & $e^{-x_1}x_1^3cos(x_1)sin(x_1)(cos(x_1)sin(x_1)^2-1)(x_2-5)$ & U$(-1, 1, 20)$ & \NguyenXOne \\
Vladislavleva-4 & $\frac{10}{5+(x1-3)^2+(x_2-3)^2+(x_3-3)^2+(x_4-3)^2+(x_5-3)^2}$ & U$(0, 2, 20)$ & \NguyenXYZDFCZ \\
Vladislavleva-5 & $30(x_1-1)\frac{x_3-1}{(x_1-10)}x_2^2$ & U$(-1, 1, 100)$ & \NguyenXYC \\
Vladislavleva-6 & $6sin(x_1)cos(x_2)$ & E$(1, 50, 50)$ & \NguyenXYC \\
Vladislavleva-7 & $2 - 2.1\cos(9.8x)\sin(1.3x_2)$ & E$(-50, 50, 10^5)$ & \NguyenXYC \\
Vladislavleva-8 & $\frac{e^{-(x-1)^2}}{1.2 + (x_2-2.5)^2}$ & U$(0.3, 4, 100)$ & \VladislavlevaB \\
\toprule
Test-2 & $3.14*x1*x1$ & U$(-1, 1, 20)$ & \NguyenXC \\
Const-Test-1 & $5*x1*x1$ & U$(-1, 1, 20)$ & \NguyenXC \\
GrammarVAE-1 & $1./3+x1+sin(x_1^2))$ & U$(-1, 1, 20)$ & \NguyenXC \\
Sine & $sin(x_1)+sin(x_1+x_1^2))$ & U$(-1, 1, 20)$ & \NguyenX \\
Nonic & $x_1^9+x_1^8+x_1^7+x_1^6+x_1^5+x_1^4+x_1^3+x_1^2+x_1$ & U$(-1, 1, 100)$ & \NguyenXY \\
Pagie-1 & $\frac{1}{1+x_1^{-4}+\frac{1}{1+x2^{-4}}} $ & E$(1, 50, 50)$ & \NguyenXY \\
Meier-3 & $\frac{x_1^2  x_2^2}{(x_1+x_2)}$ & E$(-50, 50, 10^5)$ & \NguyenXY \\
Meier-4 & $\frac{x_1^5}{x_2^3}$ & U$(0.3, 4, 100)$ & \VladislavlevaB \\
Poly-10 & $x_1x_2+x_3x4+x_5x_6+x_1x_7x_9+x_3x_6x_{10}$ & E$(-1, 1, 100)$ & \NguyenXall \\
\newline
\end{tabular}
\end{scriptsize}
\label{a-tab3}
\end{table*}

\section{Appendix: $R^2$ of SR-GPT on the AI Feynman dataset.}
We tested the performance of our proposed symbol regression algorithm, SR-GPT, on the AI Feynman dataset. This dataset contains problems from physics and mathematics across multiple subfields, such as mechanics, thermodynamics, and electromagnetism. The authors provided 100,000 sampled data points in the AI Feynman dataset, however, to better test the performance of SR-GPT, we randomly selected only 100 data points from the 100,000 provided as our experimental data. We applied SR-GPT to perform symbol regression on each data in the dataset. and recorded the $R^2$ between the predicted results and the correct answers. The experimental results indicate that SR-GPT can accurately fit the corresponding expressions from a small number of sample points. For the majority of the formulas, the $R^2$ exceeds 0.99. This indicates that the model performs well on problems in fields such as physics and mathematics, and has great potential for wide application. The experimental results are shown in Table \ref{EqTableA} and Table \ref{EqTableB}.
\begin{table*}[htp]
\centering
{\footnotesize
\begin{tabular}{|l|l|r|}
\hline
Feynman   & Equation & $R^2$ \\
\hline                            
I.6.20a       & $f = e^{-\theta^2/2}/\sqrt{2\pi}$ & 0.9992  \\
I.6.20        & $f = e^{-\frac{\theta^2}{2\sigma^2}}/\sqrt{2\pi\sigma^2}$ & 0.9988\\
I.6.20b       & $f = e^{-\frac{(\theta-\theta_1)^2}{2\sigma^2}}/\sqrt{2\pi\sigma^2}$ & 0.9923 \\
I.8.14       & $d = \sqrt{(x_2-x_1)^2+(y_2-y_1)^2}$ & 0.8929  \\
I.9.18       & $F = \frac{Gm_1m_2}{(x_2-x_1)^2+(y_2-y_1)^2+(z_2-z_1)^2}$  & 0.9944\\
I.10.7       & $F = \frac{Gm_1m_2}{(x_2-x_1)^2+(y_2-y_1)^2+(z_2-z_1)^2}$  & 0.9906\\
I.11.19      & $A = x_1y_1+x_2y_2+x_3y_3$ & 1.0   \\
I.12.1       & $F = \mu N_n$ & 1.0 \\
I.12.2       & $F = \frac{q_1q_2}{4\pi\epsilon r^2}$   & 1.0 \\
I.12.4       & $E_f = \frac{q_1}{4\pi\epsilon r^2}$  & 0.9994 \\
I.12.5       & $F = q_2 E_f$ & 1.0  \\
I.12.11      & $F = \mathcal{Q}(E_f+B v \sin\theta)$  & 0.9999 \\
I.13.4      & $K = \frac{1}{2}m(v^2+u^2+w^2)$  & 0.9969  \\
I.13.12      & $U = Gm_1m_2(\frac{1}{r_2}-\frac{1}{r_1})$ & 1.0  \\
I.14.3       & $U = mgz$ &1.0    \\
I.14.4       & $U = \frac{k_{spring}x^2}{2}$  & 0.9999  \\
I.15.3x      & $x_1 = \frac{x-ut}{\sqrt{1-u^2/c^2}}$ & 0.9993 \\
I.15.3t      & $t_1 = \frac{t-ux/c^2}{\sqrt{1-u^2/c^2}}$ & 0.9844  \\
I.15.10       & $p = \frac{m_0v}{\sqrt{1-v^2/c^2}}$ & 0.9978 \\
I.16.6       & $v_1 = \frac{u+v}{1+uv/c^2}$ & 0.9873  \\
I.18.4       & $r = \frac{m_1r_1+m_2r_2}{m_1+m_2}$ & 0.9894 \\
I.18.12      & $\tau = rF\sin\theta$  & 0.9999  \\
I.18.16      & $L = mrv \sin\theta$  & 0.9999 \\
I.24.6 & $E = \frac{1}{4} m (\omega^2+\omega_0^2) x^2$      & 0.9986\\
I.25.13      & $V_e = \frac{q}{C}$ & 1.0 \\
I.26.2       & $\theta_1 = \arcsin(n  \sin\theta_2)$ & 0.9991 \\
I.27.6       & $f_f$    $ = \frac{1}{\frac{1}{d_1}+\frac{n}{d_2}}$  & 0.9995 \\
I.29.4       & $k = \frac{\omega}{c}$ & 1.0 \\
I.29.16      & $x = \sqrt{x_1^2+x_2^2-2x_1x_2\cos(\theta_1-\theta_2)}$ & 0.9942  \\
I.30.3 & $I_* = I_{*_0}\frac{\sin^2(n\theta/2)}{\sin^2(\theta/2)}$ & 0.9912 \\
I.30.5       & $\theta = \arcsin(\frac{\lambda}{nd})$  & 0.9994\\
I.32.5       & $P = \frac{q^2a^2}{6\pi\epsilon c^3}$       & 0.9857 \\
I.32.17 & $P = (\frac{1}{2}\epsilon c E_f^2)(8\pi r^2/3) (\omega^4/(\omega^2-\omega_0^2)^2)$      & 0.9788  \\
I.34.8       & $\omega = \frac{qvB}{p}$   & 1.0\\
I.34.10       & $\omega = \frac{\omega_0}{1-v/c}$ & 0.9928 \\
I.34.14      & $\omega = \frac{1+v/c}{\sqrt{1-v^2/c^2}}\omega_0$  & 0.9992 \\
I.34.27      & $E = \hbar\omega$  & 1.0 \\
I.37.4       & $I_* = I_1+I_2+2\sqrt{I_1I_2}\cos\delta$ & 0.9927\\
I.38.12      & $r = \frac{4\pi\epsilon\hbar^2}{mq^2}$   & 0.9999  \\
I.39.10       & $E = \frac{3}{2}p_F V$     & 1.0 \\
I.39.11      & $E = \frac{1}{\gamma-1}p_F V$  & 0.9998 \\
I.39.22      & $P_F = \frac{n k_b T}{V}$       & 0.9999  \\
I.40.1       & $n = n_0e^{-\frac{mgx}{k_bT}}$    & 0.9947 \\
I.41.16      & $L_{rad} = \frac{\hbar\omega^3}{\pi^2c^2(e^{\frac{\hbar\omega}{k_bT}}-1)}$ & 0.8462  \\
I.43.16      & $v = \frac{\mu_{drift}q V_e}{d}$   & 1.0  \\
I.43.31      & $D = \mu_e k_bT$    & 1.0  \\
I.43.43      & $\kappa = \frac{1}{\gamma-1}\frac{k_bv}{A}$  & 0.9428  \\
I.44.4       & $E = n k_b T \ln(\frac{V_2}{V_1})$   & 0.8322  \\
I.47.23      & $c = \sqrt{\frac{\gamma pr}{\rho}}$   & 0.9926\\
I.48.20       & $E = \frac{m c^2}{\sqrt{1-v^2/c^2}}$ &  0.8859\\
I.50.26 & $x = x_1[\cos(\omega t)+\alpha\> cos(\omega t)^2]$      & 0.9999   \\
\hline
\end{tabular}
\caption{Tested Feynman Equations, part 1. }
\label{EqTableA}
}
\end{table*}

\begin{table*}[htp]
\centering
{\footnotesize
\begin{tabular}{|l|l|r|}
\hline
Feynman   & Equation & $R^2$\\
\hline       
II.2.42   & P     $ = \frac{\kappa(T_2-T_1)A}{d}$  & 0.7842  \\
II.3.24   & $F_E = \frac{P}{4\pi r^2}$  & 0.9976 \\
II.4.23   & $V_e = \frac{q}{4\pi\epsilon r}$   & 0.9997 \\
II.6.11 & $V_e =\frac{1}{4\pi\epsilon}\frac{p_d\cos \theta}{r^2}$      & 1.0 \\
II.6.15a & $E_f = \frac{3}{4\pi\epsilon}\frac{p_d z}{r^5} \sqrt{x^2+y^2}$      & 0.9466  \\
II.6.15b & $E_f = \frac{3}{4\pi\epsilon}\frac{p_d}{r^3} \cos\theta\sin\theta$      & 0.9943  \\
II.8.7    & $E = \frac{3}{5}\frac{q^2}{4\pi\epsilon d}$  & 0.9955  \\
II.8.31   & $E_{den} = \frac{\epsilon E_f^2}{2}$                     & 1.0 \\
II.10.9   & $E_f = \frac{\sigma_{den}}{\epsilon}\frac{1}{1+\chi}$      & 0.9999  \\
II.11.3 & $x = \frac{q E_f}{m(\omega_0^2-\omega^2)}$      & 0.9901     \\
II.11.7 & $n = n_0(1+ \frac{p_d E_f \cos\theta}{k_b T})$      & 0.8826 \\
II.11.20  & $P_* = \frac{n_\rho p_d^2 E_f}{3 k_b T}$ & 0.7783  \\
II.11.27 & $P_* = \frac{n\alpha}{1-n\alpha/3}\epsilon E_f$      & 0.9859   \\
II.11.28  & $\theta = 1+\frac{n\alpha}{1-(n\alpha/3)}$    & 0.9947\\ 
II.13.17  & $B = \frac{1}{4 \pi \epsilon c^2}\frac{2I}{r}$ & 0.9997\\
II.13.23  & $\rho_c = \frac{\rho_{c_0}}{\sqrt{1-v^2/c^2}}$          & 0.9807  \\
II.13.34  & $j = \frac{\rho_{c_0}v}{\sqrt{1-v^2/c^2}}$     & 0.9938 \\
II.15.4   & $E = -\mu_M B \cos\theta$               & 1.0 \\
II.15.5   & $E = -p_d E_f\cos\theta$  & 1.0 \\
II.21.32  & $V_e = \frac{q}{4\pi\epsilon r(1-v/c)}$   & 0.9954   \\
II.24.17 & $k = \sqrt{\frac{\omega^2}{c^2}-\frac{\pi^2}{d^2}}$      & 0.9872   \\
II.27.16  & $F_E = \epsilon c E_f^2$        & 1.0 \\
II.27.18  & $E_{den} = \epsilon E_f^2$         & 1.0 \\
II.34.2a  & $I = \frac{qv}{2\pi r}$         & 0.9982 \\
II.34.2   & $\mu_M = \frac{q v r}{2}$             & 0.9918 \\
II.34.11  & $\omega = \frac{g_{\_} q B}{2m}$          & 0.9937 \\
II.34.29a & $\mu_M = \frac{q h}{4\pi m}$      & 1.0  \\
II.34.29b & $E = \frac{g_{\_} \mu_M B J_z}{\hbar}$ & 0.8882\\
II.35.18 & $n = \frac{n_0}{\exp(\mu_m B/(k_b T))+\exp(-\mu_m B/(k_b T))}$      & 0.9466 \\
II.35.21  & $M = n_\rho \mu_M \tanh(\frac{\mu_M B}{k_b T})$     & 0.8722 \\
II.36.38 & $f = \frac{\mu_m B}{k_b T}+\frac{\mu_m\alpha M}{\epsilon c^2 k_b T}$      & 0.9244\\
II.37.1   & $E = \mu_M(1+\chi)B$    & 0.9999\\
II.38.3   & $F = \frac{Y A x}{d}$            & 1.0 \\
II.38.14  & $\mu_S = \frac{Y}{2(1+\sigma)}$     & 0.9999  \\
III.4.32  & $n = \frac{1}{e^{\frac{\hbar\omega}{k_bT}}-1}$ & 0.9877  \\
III.4.33  & $E = \frac{\hbar\omega}{e^{\frac{\hbar\omega}{k_b T}}-1}$  & 0.9998    \\
III.7.38  & $\omega = \frac{2 \mu_M B}{\hbar}$  & 0.9914  \\
III.8.54  & $p_{\gamma}$    $ = \sin(\frac{E t}{\hbar})^2$  & 0.9943\\
III.9.52  & $p_{\gamma}$    $ = \frac{p_d E_f t}{\hbar} \frac{    \sin((\omega-\omega_0)t/2)^2}{((\omega-\omega_0)t/2)^2}$ & 0.7266  \\
III.10.19 & $E = \mu_M\sqrt{B_x^2+B_y^2+B_z^2}$  & 0.9928 \\
III.12.43 & $L = n\hbar$ & 1.0  \\
III.13.18 & $v = \frac{2 E d^2 k}{\hbar}$ & 0.9999  \\
III.14.14 & $I = I_0 (e^{\frac{q V_e}{k_b T}}-1)$  & 0.9982\\
III.15.12 & $E = 2U(1-\cos(kd))$    & 0.9999 \\
III.15.14 & $m = \frac{\hbar^2}{2E d^2}$     & 0.9983  \\
III.15.27 & $k = \frac{2\pi\alpha}{nd}$    & 0.9998 \\
III.17.37 & $f = \beta(1+\alpha \cos\theta)$ & 1.0 \\
III.19.51 & $E = \frac{-mq^4}{2(4\pi\epsilon)^2\hbar^2}\frac{1}{n^2}$     & 0.9894 \\
III.21.20 & $j = \frac{-\rho_{c_0} q A_{vec}}{m}$  & 0.7489  \\
\hline
\end{tabular} 
\caption{Tested Feynman Equations, part 2. }
\label{EqTableB}

}
\end{table*}
\section{Appendix: Related work supplement} 
\textbf{Self-Learning\_Gene\_Expression\_Programming (SL-GEP)}\citep{gp2}, The SL-GEP method utilizes Gene Expression Programming (GEP) to represent each chromosome, which consists of a main program and a set of Automatically Defined Functions (ADFs). Each ADF is a sub-function used to solve sub-problems and is combined with the main program to address the target problem of interest. In the initialization step, all encoded ADFs in each chromosome are randomly generated. Then, during the evolutionary search process, SL-GEP employs a self-learning mechanism to improve the search outcomes. Specifically, SL-GEP utilizes an adaptive learning algorithm to dynamically evolve the ADFs online and integrate them with the main program to construct more complex and higher-order sub-functions, thereby enhancing search accuracy and efficiency.\\
\textbf{semantic genetic programming (SGD)}\citep{gp3}, Traditional genetic programming approaches often rely on random search to find optimal solutions, but this method is inefficient and prone to getting stuck in local optima. Therefore, SGD utilizes program behavior to guide the search, aiming to improve the efficiency and accuracy of symbolic regression problems. Specifically, this method starts by transforming input data into vector form and uses it as a constraint in a linear programming model. Then, semantic information is employed to evaluate each program and classify them based on their behavioral characteristics. Subsequently, the best programs are selected within each category and used to generate a new generation of programs. This approach effectively reduces the search space and accelerates convergence speed.\\
\textbf{shape-constrained symbolic regression (SCSR) }\citep{gp4}, The main idea of SCSR is a shape-constrained symbolic regression algorithm. This method leverages prior knowledge about the shape of the regression function to improve the accuracy of the regression model. Specifically, the article introduces both single-objective and multi-objective algorithms. The single-objective algorithm utilizes genetic programming techniques to generate the best-fitting curve. On the other hand, the multi-objective algorithm considers multiple optimization objectives and employs Pareto front techniques to search for a set of non-dominated solutions.
\section{Appendix:  Computing resources} 
The server we use is equipped with an Intel(R) Xeon(R) Gold 5218R CPU, which has a base frequency of 2.10 GHz. It has a total of 20 CPU cores, allowing for parallel processing and improved computational performance. The high core count and efficient architecture of the Intel Xeon Gold 5218R make it suitable for handling demanding computational tasks and workloads.
\newpage
\section{Appendix:  MCTS} \label{sec:MCTS}
In order to clearly show the MCTS search process, we assume that there are only two basic symbols [sin,x]. The target expression is y = sin(x). The search process is as follows.

\textbf{Initialization:} Initially there is a root node $S_0$, and each node in the tree has two values, the reward Q of the node and the number of visits to that node N.\\
\begin{center}
\includegraphics[width=100mm]{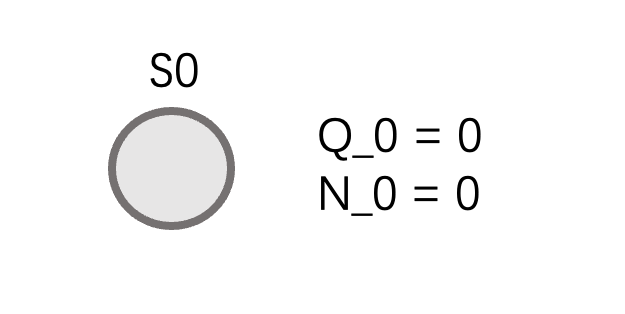}
\end{center}
\textbf{First iteration:} Node $S_0$ is the root and leaf node, and is not the terminating node, so it is extended. Assume that $S_0$ has two actions (the basic symbol [sin,x]) after it , which are transferred to $S_1$ and $S_2$ respectively.
\begin{center}
\includegraphics[width=100mm]{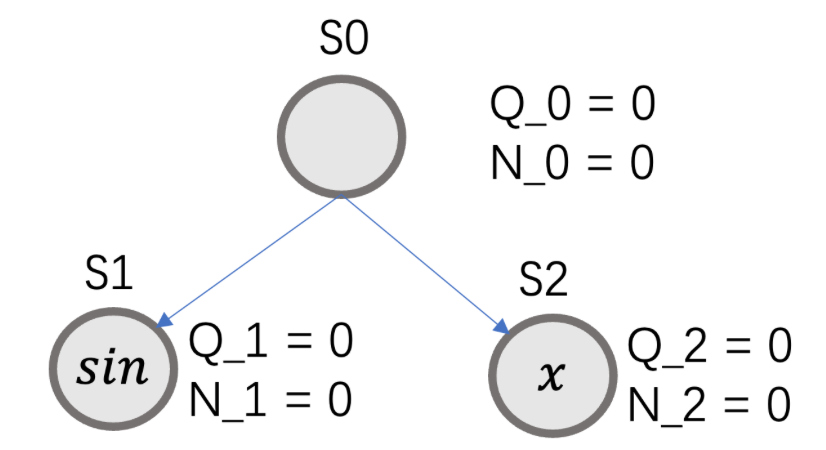}
\end{center}

You can then use the UCT formula to choose whether to extend $S_1$or $S_2$. Here $N_1$and $N_2$are both 0, so the UCT value of both nodes is infinite, so any node can be selected, here $S_1$is selected for extension and simulation (random selection of symbols). After simulation, it was found that the final reward value was 0.2, so it was updated retrospectively. $Q_1= 0.2$,$N_1$= 1,$Q_0$=0.2,$N_0$= 1.\\
\begin{center}
\hspace{-0.535cm}\includegraphics[width=100mm]{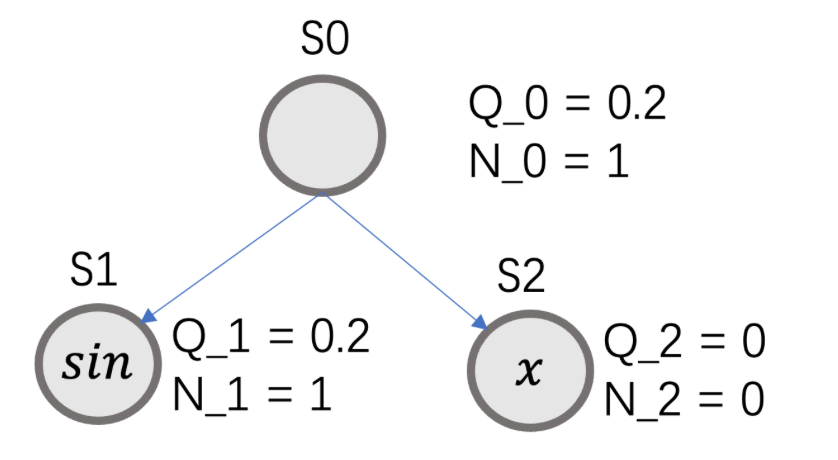}
\end{center}
\newpage
\textbf{The second iteration:} Starting from $S_0$, calculate the UCT values of $S_1$ and $S_2$, and select the larger one for expansion. (assuming $S_1>S_2$ after calculation)\\
Then according to the UCT value, $S_1$ is selected for expansion. After reaching $S_1$, it is found that it is a leaf node and has been explored, then enumerate all possible states of the current node (each action corresponds to a state), and add them to the tree.

\begin{center}
\hspace{-0.535cm}\includegraphics[width=100mm]{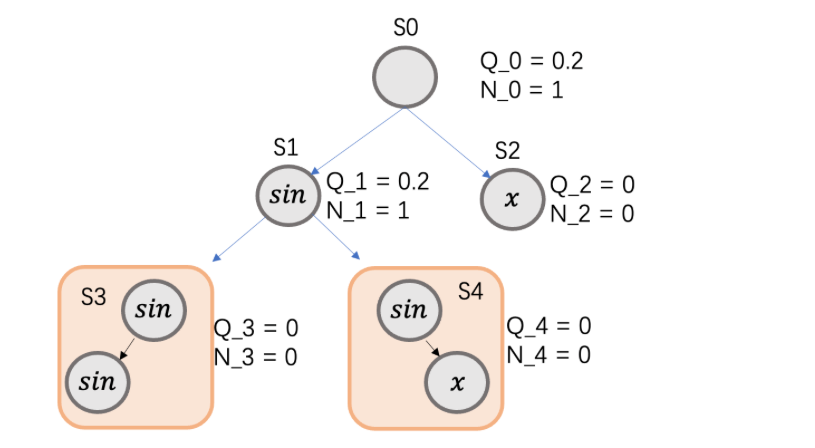}
\end{center}
Then we can select either $S_3$ or $S_4$ at random as before. Keep iterating. (In this example, S4 has successfully found the target expression)

\end{document}